\documentclass[11pt,a4paper]{article}

\PassOptionsToPackage{expansion=false}{microtype}

\usepackage{pasteurlabs}

\usepackage{makecell}

\newcommand{\rwp}{RWP}
\newcommand{\rwr}{RWR}

\title{Read, Write, Relax: Why Neural PDE Surrogates Need Both Global and Local Processing}

\author{Anuj Kumar\textsuperscript{1},\;
        Heiko Zimmermann\textsuperscript{1},\;
        Josiah Bjorgaard\textsuperscript{1},\;
        Jacan Chaplais\textsuperscript{1},\;
        Nikolaos Bouklas\textsuperscript{1,2},\;
        Matteo Salvador\textsuperscript{1},\;
        Alexander Lavin\textsuperscript{1,3}}

\date{August 2026}

\affiliation{%
  \textsuperscript{1}Pasteur Labs, Brooklyn, NY, USA\quad
  \textsuperscript{2}Cornell University, Ithaca, NY, USA\quad
  \textsuperscript{3}Institute for Simulation Intelligence, New York, NY, USA}

\correspondence{lavin@simulation.science}

\organization{\includegraphics[height=26pt]{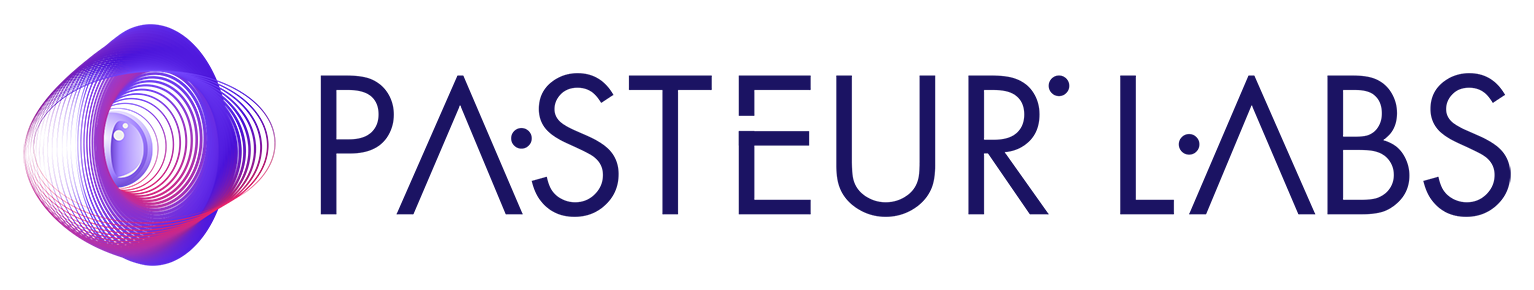}}

\shorttitle{Read, Write, Relax: Why Neural PDE Surrogates Need Both Global and Local Processing}

\begin{document}
\sloppy
% Keep floats near their mention in single-column layout, with tighter float
% and caption spacing to reduce whitespace.
\setlength{\textfloatsep}{13pt plus 2pt minus 4pt}
\setlength{\intextsep}{11pt plus 2pt minus 2pt}
\setlength{\floatsep}{10pt plus 2pt minus 2pt}
\captionsetup{skip=6pt}
\renewcommand{\topfraction}{0.85}
\renewcommand{\bottomfraction}{0.6}
\renewcommand{\textfraction}{0.1}
\renewcommand{\floatpagefraction}{0.75}
\maketitle

\begin{plabstract}
Recent mesh-based simulation advances have, in no small part, relied on neural surrogates of two distinct families: global models that route information through a small set of latent tokens, and local models that perform message passing across mesh edges. Consistent with both classes is the inability to perform beyond low-dimensional problems and small-scale or oversimplified meshes, the simulation regimes where industrial problems reside. Our work shows this explicitly and presents a unified formulation. In global approaches, latent-token attention acts as a spatial low-pass filter, while local message passing lacks the global reach necessary to propagate information across large mesh spaces. Viewed through the error, the two operators are the halves of a multigrid cycle: one corrects errors at the lower end of the spectrum, the other at the higher end, and neither can do the other's job. We introduce Read-Write-Relax (RWR), which interleaves latent attention with message-passing relaxation under a unified formulation. The interleaved processor lowers error across the entire spectrum, making RWR the most accurate model in nearly every comparison across our industrial and public benchmarks. It is also markedly data-efficient in the scarce-data regimes, accurate on the engineering quantities of interest, and scales full-field predictions to challenging, large-scale problems.
\end{plabstract}

% ─────────────────────────────────────────────────────────────────────────────
\section{Introduction}
\label{sec:intro}

Learned surrogates for mesh-based simulation have converged on two designs. Global models compress the discretized field into a small set of latent tokens, process the tokens, and expand the result back onto the mesh, a pattern shared by Perceiver-style encoders~\cite{jaegle2021perceiver,jaegle2021perceiverio}, Transolver's physics attention~\cite{wu2024transolver}, and FLARE's low-rank routing~\cite{puri2025flare}; we refer to these global models as \emph{latent-attention} models. Local models such as MeshGraphNets (MGN)~\cite{pfaff2020mgn} update each node from its mesh neighbors, realizing learnable stencils that resolve structure down to the mesh scale but communicate only within a receptive field that grows by one hop per layer~\cite{alon2021on}.

Steady-state problems stress both designs at once. At convergence, the solution everywhere is shaped by the boundary and operating conditions together with the geometric parameterization, so a surrogate must propagate information across the entire domain in a single forward pass rather than over many rollout steps, while simultaneously resolving the steep local gradients of the solution. Global communication and local resolution are therefore both required, and each family delivers one while falling short on the other. The local half carries particular weight in practice, because the outputs that engineering decisions rest on are derivative-based, among them wall shear stress, vorticity, and integrated losses computed from them; we call them \emph{engineering quantities}.

The way surrogates for Computer-Aided Engineering (CAE) are trained and evaluated magnifies both shortfalls, the local family's limited communication and the global family's limited resolution. On the training side, every sample is the output of an expensive solver run, with industrial-grade Reynolds-Averaged Navier--Stokes (RANS) and hybrid RANS--large-eddy simulations costing hundreds to thousands of core-hours per geometry, so CAE datasets number in the hundreds of samples rather than the millions common elsewhere in deep learning~\cite{bonnet2022airfrans,ashton2024ahmedml,ashton2024drivaerml}. Data efficiency is therefore a requirement rather than a convenience. On the evaluation side, progress is predominantly reported on bulk metrics such as the Mean Absolute Error (MAE), relative $L_2$ error, and $R^2$, and these are documented to be poor proxies for the engineering quantities just defined~\cite{takamoto2022pdebench,bonnet2022airfrans}.

We take up these requirements on two fronts, an analysis of why each family fails and an architecture built on the answer. The analysis compares a message-passing model, MGN, with a latent-attention model we call the Read-Write Perceiver (\rwp{}), tracking their error spectra over the course of training. The two designs fail from opposite ends of the spectrum, and read through multigrid~\cite{brandt1977multi,briggs2000multigrid}, they occupy the two complementary roles of a cycle. Read-Write-Relax (\rwr{}), our architecture, interleaves the two mechanisms and takes its name from this reading. The read and the write act as the restriction and prolongation of a coarse correction in latent space, and the message-passing sweeps act as the learned relaxation. The combined processor recovers the entire spectrum, carries the accuracy into the gradients and wall quantities that make up the engineering quantities, and holds where it matters most in practice, at small training budgets, where the latent-attention baselines degrade fastest.

\paragraph{Contributions}
\begin{enumerate}
\item \textbf{The analysis.} We place recent global models in one family, Bottlenecked Self-Attention (BSA), of which our \rwp{} is itself a member, and characterize MGN, \rwp{}, and \rwr{} under a single spectral lens on an industrial steady-flow benchmark. This yields a unified analysis of their complementary failure modes across the error spectrum (Sec.~\ref{sec:experiments}).
\item \textbf{The architecture.} \rwr{} interleaves latent attention with message-passing relaxation under a unified configuration $(m_1, r, m_2)$ that prescribes how much local and global processing the model performs, with MGN and \rwp{} recovered as special cases. Geometry-anchored queries and boundary-only encoding, which lets the attention read only the information-dense boundary points, keep the design scalable to large industrial problems, and the resulting model is markedly data-efficient, delivering accurate engineering quantities from far fewer training samples than the latent-attention baselines. In addition, across the benchmarks, \rwr{} is the most accurate model in nearly every comparison (Secs.~\ref{sec:model}, \ref{sec:experiments}).
\end{enumerate}

% ─────────────────────────────────────────────────────────────────────────────
\section{Related Work}
\label{sec:related}

\paragraph{Attention and message passing on graphs}
The conflict between local message passing and global information exchange is well documented in graph learning. Message passing cannot reach beyond its receptive field, and even within it, information from exponentially growing neighborhoods is compressed into fixed-size vectors~\cite{alon2021on}, a failure mode known as over-squashing, later linked to negatively curved edges in the graph~\cite{topping2022understanding}. On the other hand, graph transformers discard message passing altogether and attend over all node pairs, relying on spectral or structural encodings to reinstate the topology that attention alone would ignore~\cite{kreuzer2021rethinking,ying2021transformers}. At the opposite and cheapest end, a single virtual node grafted onto a message-passing network acts as a global scratch space~\cite{gilmer2017neural}, a mechanism that provably approximates linear attention~\cite{cai2023connection}. Hybrid approaches define the middle ground, where some run a Graph Neural Network (GNN) first and a transformer afterward, so that local structure is summarized before global reasoning~\cite{wu2021representing}, while others run message passing and global attention in parallel within every layer~\cite{rampasek2022recipe,shirzad2023exphormer}. Our architecture follows this lineage but alternates the two mechanisms across the depth of the network, with the global pathway realized as cross-attention to a small latent set rather than attention over all nodes.

\paragraph{Local and global processing in PDE surrogates}
The same problem arises in mesh-based PDE surrogates. MGN-style processors~\cite{pfaff2020mgn} need message-passing depth proportional to the mesh diameter, which is particularly limiting for steady and elliptic problems where boundary information influences the interior globally~\cite{wang2024beno} and accurate prediction requires long-range exchange that deep stacks capture poorly and train slowly~\cite{ripken2023multiscale,gladstone2024mesh}. The same issue appears in time-dependent regimes: parabolic problems require stable long-horizon propagation, and hyperbolic problems require sharp, non-dissipative transport of features like shock waves, both hard for a processor whose receptive field scales only with depth. Early responses built multiscale hierarchies~\cite{fortunato2022multiscale,cao2023efficient}, and EAGLE introduced global attention over a pooled, clustered mesh~\cite{janny2023eagle}. A second family adopts a stacked composition that aggregates locally once and then processes globally with transformers, as in GINO~\cite{li2023gino} and GAOT~\cite{wen2025gaot}. A third line of work keeps both mechanisms inside the processor. GITO and FRGT combine the two mechanisms in parallel and stacked form, respectively~\cite{ramezankhani2025gito,duthe2025graph}. Most relevant is MeshTransolver, which wraps a Transolver-style token-attention core between message-passing pre- and post-processing stages~\cite{curtosi2026crash}, a sandwich composition that concurrent work adopts as well~\cite{iparraguirre2026mgnt}. These designs fix the composition in advance as parallel, stacked, or a single wrapped core, and are validated by end-task accuracy, largely leaving open why the combination outperforms either mechanism alone. \rwr{} alternates relaxation and latent-attention iterations throughout the network and exposes the mix as a configuration, and the analysis of Sec.~\ref{sec:experiments} supplies the missing account in spectral terms.

\paragraph{Latent-token surrogates}
A parallel line of latent-token surrogates, including Perceiver IO~\cite{jaegle2021perceiverio}, Transolver and its successors~\cite{wu2024transolver,luo2025transolverpp,adams2025geotransolver}, LNO~\cite{wang2024latent}, UPT~\cite{alkin2024upt}, AROMA~\cite{serrano2024aroma}, and FLARE~\cite{puri2025flare}, compresses the field into a small set of tokens and processes it there. We return to these models in the analysis (Sec.~\ref{sec:experiments}), where we show that they instantiate a single BSA template together with our own \rwp{}.

% ─────────────────────────────────────────────────────────────────────────────
\section{The Model: Read-Write-Relax}
\label{sec:model}

\begin{figure}[htbp]
\centering
\includegraphics[width=0.7\linewidth]{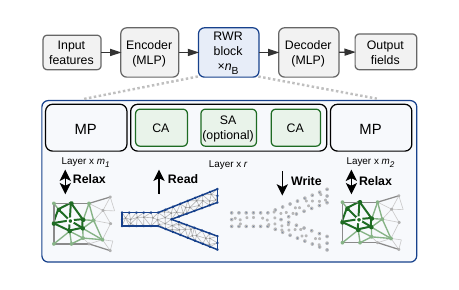}\\[2pt]
\caption{One \rwr{} block. The encoder and decoder are pointwise MLPs; the processor runs $m_1$ message-passing sweeps (\emph{relax}), $r$ read-write iterations against $L$ persistent latent tokens, and $m_2$ further sweeps; $n_B$ such blocks are stacked. Setting $r{=}0$ recovers MGN, and setting $m_1{=}m_2{=}0$ recovers \rwp{}. MP denotes message passing, CA cross-attention, and SA self-attention.}
\label{fig:rwr}
\end{figure}

Our surrogate follows the encode-process-decode organization of MGN~\cite{pfaff2020mgn}: node and edge features are lifted by pointwise encoders, transformed by a processor, and projected to output fields by a decoder. \rwr{} is the processor. It consists of $n_B$ blocks, each parameterized by a configuration $(m_1, r, m_2)$ that runs $m_1$ message-passing sweeps, $r$ read-write iterations against a set of $L$ latent tokens, and $m_2$ further sweeps (Fig.~\ref{fig:rwr}). The configuration prescribes how much local and global processing the model performs. In particular, $r = 0$ leaves only message passing and recovers MGN, while setting $m_1 = m_2 = 0$ removes the relaxation sweeps entirely and yields the \rwp{}, the latent-attention model we analyze alongside \rwr{} throughout the paper. All three models in our experiments are therefore the same implementation at different configurations.

Let $U \in \mathbb{R}^{N \times C}$ denote the node features, $P$ their positional encodings, and $Z \in \mathbb{R}^{L \times C}$ the latents. One read-write iteration applies
\begin{equation}
\begin{aligned}
\textbf{read:} \quad & Z \leftarrow Z + \mathrm{CA}\!\left(Q{=}Z,\; KV{=}U|_{\mathcal{R}} + P|_{\mathcal{R}}\right) \\
\textbf{write:} \quad & U \leftarrow U + \mathrm{CA}\!\left(Q{=}U + P,\; KV{=}Z\right),
\end{aligned}
\label{eq:rwp}
\end{equation}
where $\mathrm{CA}$ is pre-norm cross-attention with a feed-forward layer and $\mathcal{R}$ is the read set. An optional latent self-attention tower can be inserted between the read and the write. The latents persist across the $r$ iterations of a block, so each read refines the state accumulated by earlier ones rather than rebuilding it, a design shared with the read-process-write generator of \citet{jabri2023rin}. The relaxation sweeps are standard message-passing updates with residual connections, in which edge features are updated from their endpoint nodes, and node features are updated from their aggregated incident edges~\cite{pfaff2020mgn}. The names record the multigrid reading introduced above. The read acts as a restriction onto an $L$-dimensional coarse space, the write as a prolongation back to the mesh, and together with the latent update they form a coarse correction; the message-passing sweeps act as learned relaxation.

\paragraph{Geometry-anchored queries}
The latent queries are anchored in the geometry rather than learned as free vectors. Each latent possesses a learnable anchor coordinate in physical space, and its query is produced by passing this coordinate through the same random-Fourier-feature embedding~\cite{tancik2020rff} that encodes node positions for the read and the write, so queries and keys meet in a single coordinate space and each latent costs three parameters instead of a full feature vector. One embedding is shared across all blocks, which ties the frequency content of the positional encoding across the processor.

\paragraph{Boundary-only encoding}
\label{sec:boundary-read}
Because the read and the write are separate attention maps, the read set $\mathcal{R}$ need not cover the full mesh. For steady problems we can optionally restrict it to the boundary points, where the geometry, the boundary conditions, and hence the identity of the problem are concentrated. The choice has a classical justification, since for elliptic problems the interior solution is determined by boundary data through the Green representation~\cite{kress2014linear}, an observation that boundary-embedded neural operators exploit directly~\cite{wang2024beno}, and it reduces the read cost from $O(NL)$ to $O(N_b L)$ with $N_b \ll N$ boundary points. The write continues to address every node, since the output lives on the full mesh.

\section{Datasets}
\label{sec:datasets}

\begin{figure}[htbp]
\centering
\begin{minipage}[c]{0.35\linewidth}\centering
\includegraphics[width=\linewidth]{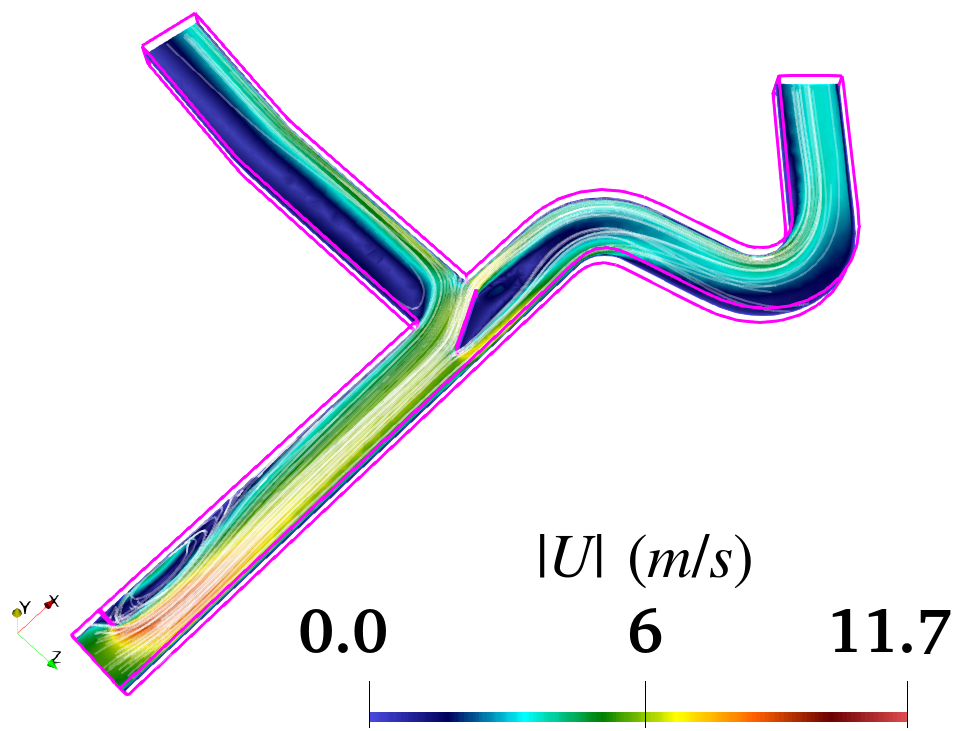}
\end{minipage}\hspace{0.05\linewidth}%
\begin{minipage}[c]{0.35\linewidth}\centering
\includegraphics[width=\linewidth]{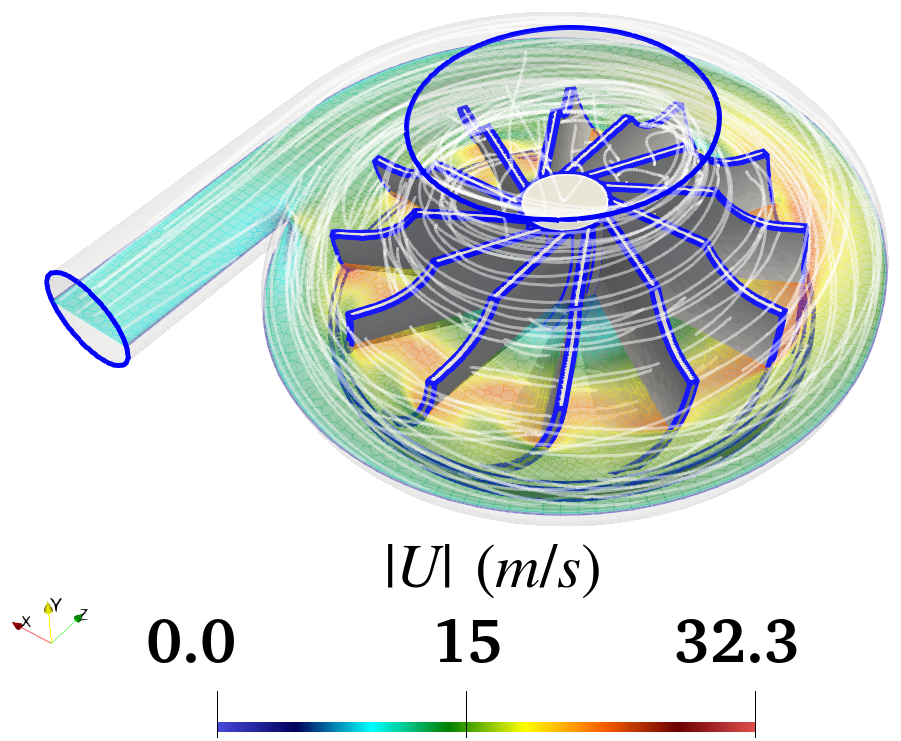}
\end{minipage}
\caption{The two simulation environments, branched pipe (left) and centrifugal pump (right), showing a representative velocity field.}
\label{fig:datasets}
\end{figure}

We evaluate the surrogate models on two industrial steady-flow problems (Fig.~\ref{fig:datasets}), the second of which we use at two extents.

\paragraph{Branched pipe} Steady flow through a Y-shaped Heating-Ventilation-Air-Conditioning (HVAC) duct junction whose per-channel flow rates are controlled by baffle plates; the sharp turning angles at the baffles make the flow at the junction highly turbulent. Samples vary the duct geometry, the baffle angles, and the inlet velocity, and the modeled fields are the pressure and the two in-plane velocity components ($p$, $u_x$, $u_z$) on meshes of roughly 60k cells. All analyses of Sec.~\ref{sec:experiments} run on this benchmark.

\paragraph{Centrifugal pump} Steady flow through a centrifugal pump, in which a rotating impeller draws fluid in through the eye of a stationary casing and discharges it radially through the volute; the rotating blades raise both the pressure and the velocity of the fluid, and the rotation is modeled with a moving reference frame. Samples vary the inlet and impeller radii, the blade pitch angle, and the inlet velocity, and the modeled fields are the three velocity components and the pressure. We use it at two extents, the casing region alone, the stationary zone of the simulation with roughly 270k cells, and the full production mesh including the rotating impeller zone with roughly 1.4M cells; both appear in the benchmarks of Sec.~\ref{sec:benchmarks}.

Both datasets are generated with a finite-volume RANS setup. Simulation configurations, design-parameter ranges, and dataset splits are given in App.~\ref{app:datasets}.

% ─────────────────────────────────────────────────────────────────────────────
\section{Results and Discussion}
\label{sec:experiments}

All analyses in this section, except for the ones in Sec.~\ref{sec:benchmarks}, use the branched pipe benchmark. MGN, \rwp{}, and \rwr{} are trained as configurations of the same \rwr{} processor at matched parameter counts and identical training setups; where the comparison concerns the latent-attention family more broadly, we also train Transolver and GeoTransolver in the same pipeline under the same budget. Errors are reported in physical space and spectrally. We use the error spectrum $E_{\mathrm{err}}(k)$, computed on a mid-plane slice of the domain, and gradient and wall metrics for the engineering quantities; band-resolved errors in the convention of PDEBench~\cite{takamoto2022pdebench} are reported in the appendix (Table~\ref{tab:full-metrics}). Precise metric definitions (App.~\ref{app:metrics}), together with model configurations and training details (App.~\ref{app:infra}), are given in the appendix as well.

\subsection{One Global Family: BSA}
\label{sec:bsa}

The ideal global mixer is full self-attention over all $N$ mesh points, which is message passing on the complete graph and couples every point to every other in a single layer. Its $O(N^2)$ cost is prohibitive at industrial mesh sizes, so the recent global models buy the coupling at a discount by routing it through $L \ll N$ latent tokens, and the discount has one common form. In the notation of Sec.~\ref{sec:model}, one layer of the family computes
\begin{equation}
Z = A\,V(U), \qquad Z' = g(Z), \qquad U \leftarrow U + B\,Z'
\label{eq:bsa}
\end{equation}
with a row-stochastic read $A \in \mathbb{R}^{L \times N}$, a value projection $V$, a latent processor $g$, and a row-stochastic write $B \in \mathbb{R}^{N \times L}$. We call this template \emph{Bottlenecked Self-Attention} (BSA). The induced point-to-point mixing matrix $W = BA$ has rank at most $L$, a bound FLARE states for its own mixing operator~\cite{puri2025flare} and which holds across the family: each layer can represent the mesh with at most $L$ spatial basis fields.

The template thus gives the family a common vocabulary, a read that compresses the field into tokens, a latent update, and a write that broadcasts the update back, and the recent global models differ only in how each of these three roles is parameterized. Table~\ref{tab:bsa-family} places them in the template. Perceiver IO reads once into latents and decodes once at the end~\cite{jaegle2021perceiverio}. Transolver's slices are latents, since its slice weights are a softmax read, its slice tokens the weighted accumulation of the points, self-attention runs over the $L$ slice tokens, and the same tied weights broadcast the result back~\cite{wu2024transolver}. LNO's physics-cross-attention projects the field onto $L$ learnable position embeddings and inverts the projection at the output~\cite{wang2024latent}. UPT aggregates the mesh onto supernodes and pools them into latent tokens that a transformer processes end to end~\cite{alkin2024upt}. AROMA encodes into $L$ tokens refined by a latent diffusion transformer, decoding through query-local neural fields~\cite{serrano2024aroma}. FLARE routes through per-head latent tokens with no latent processing at all~\cite{puri2025flare}. Our \rwp{} belongs to the same family, built for repeated use inside a processor with a decoupled read and write and latents that persist across iterations.

The kind of update this rank-bounded template permits is illustrated empirically by the training dynamics below.

\begin{table}[htbp]
\centering
\footnotesize
\setlength{\tabcolsep}{3pt}
\begin{tabular}{@{}lcccc@{}}
\toprule
 & read & process & write & write-back freq. \\
\midrule
Perceiver IO & queries & SA & out queries & once \\
Transolver family & slice weights & SA & tied weights & layer \\
LNO & learned PE & SA & inverse proj. & once \\
UPT & MP + pool & SA & perceiver & once \\
AROMA & cross-attn & DiT & query-local & once \\
FLARE & queries/head & --- & roles swapped & layer \\
\textbf{\rwp{} (ours)} & geo-anchored & SA (opt.) & decoupled & iter. \\
\bottomrule
\end{tabular}
\caption{The BSA family, Eq.~\eqref{eq:bsa}. All members route the field update through $L \ll N$ latent tokens and inherit the rank bound; they differ in how the read and write are parameterized and how often the write recurs (once at decode, every layer, or every iteration). \rwp{}'s latents persist across iterations within a block, and its read may further be restricted to the boundary points. The Transolver family row covers Transolver, Transolver++, and GeoTransolver, which share the slice read and write. SA denotes self-attention over the latent tokens, DiT a diffusion transformer, PE a positional embedding, and MP message passing; in FLARE the read and write swap the query and key-value roles between the same projections.}
\label{tab:bsa-family}
\end{table}

\begin{figure*}[htbp]
\centering
\includegraphics[width=0.98\linewidth]{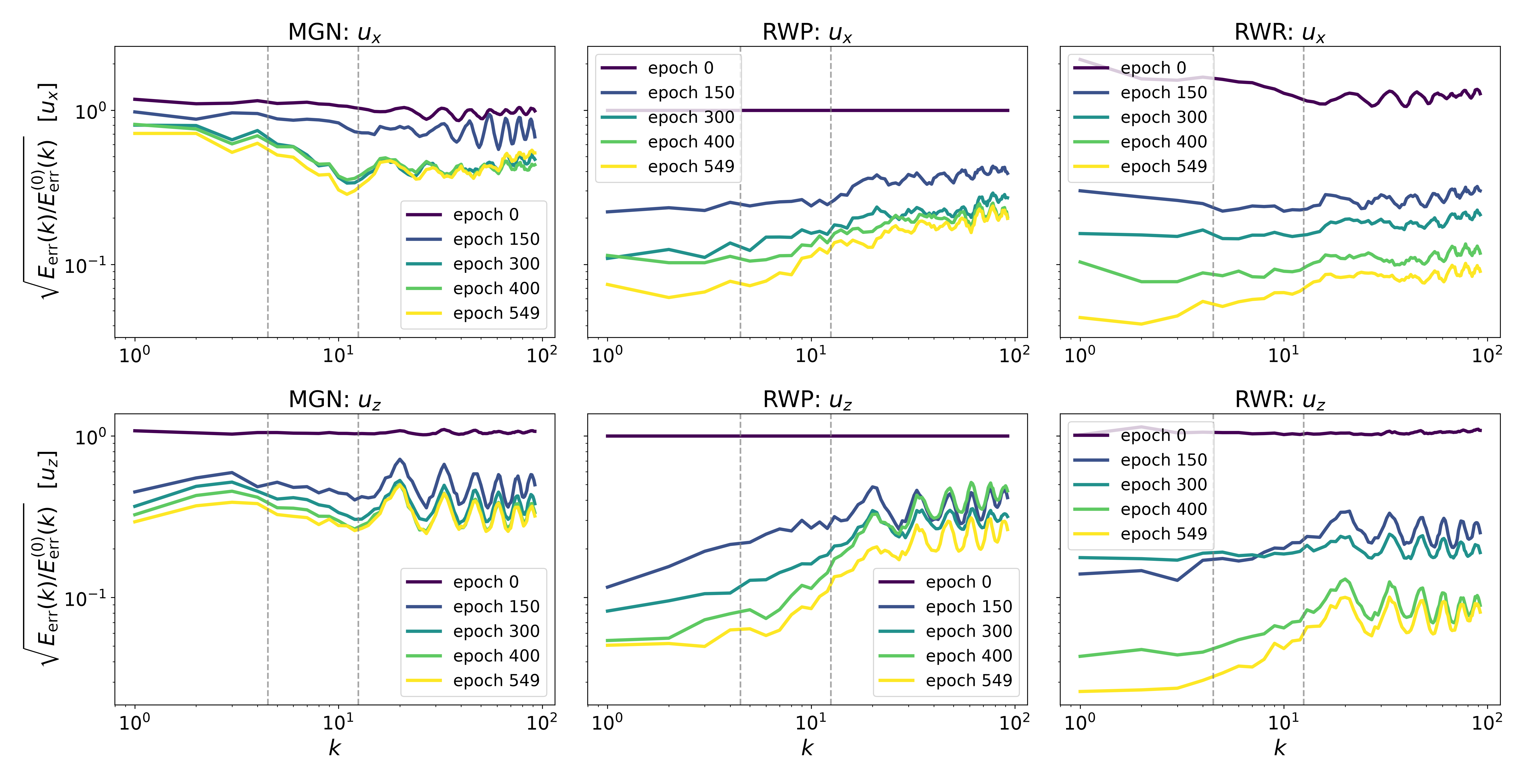}
\caption{Normalized error spectra over training on held-out branched pipe cases. Each panel shows $\sqrt{E_{\mathrm{err}}(k)/E_{\mathrm{err}}^{(0)}(k)}$, the error amplitude at wavenumber $k$ normalized by the initial error of \rwp{}, for the two velocity components (rows) under MGN, \rwp{}, and \rwr{} (columns) at five training epochs. Dashed lines mark the low/mid/high band cutoffs.}
\label{fig:training-dynamics}
\end{figure*}

\subsection{Latent Attention Behaves as a Spatial Low-Pass Filter}
\label{sec:lowpass-analysis}

Fig.~\ref{fig:training-dynamics} tracks the error spectrum of each model over the course of training. The \rwp{} column shows a strong and persistent tilt. Error in the low band contracts by more than an order of magnitude and continues to contract as training proceeds, while the mid- and high-band errors stall early and settle several times higher. Whatever the model learns, it learns overwhelmingly at large spatial scales. The latent-attention processor behaves as a spatial low-pass filter. The band-resolved training curves (Fig.~\ref{fig:training-bands} in the appendix) show the same picture, with \rwp{}'s high band separating from its low band early in training and never rejoining it.

We attribute this behavior to several compounding factors rather than to a single cause. The first is the bottleneck itself, since a BSA update spans at most $L$ spatial basis fields per layer, and covering the domain with $L$ read profiles forces every latent to average over a finite region, attenuating the scales finer than that region. The second is attention, independent of any bottleneck, since a row-stochastic attention matrix is a kernel smoother~\cite{tsai2019transformer} and provably a low-pass filter under repeated application~\cite{wang2022antioversmoothing,park2022how}. The third is the loss, since the spectra of PDE solutions decay with wavenumber, so the low wavenumbers dominate a Euclidean objective while the high wavenumbers contribute almost nothing, a documented driver of over-smoothed neural-operator predictions~\cite{lippe2023pderefiner,qin2024toward,khodakarami2025mitigating}. The fourth, not independent of the third, is spectral bias, since networks fit low frequencies before high ones~\cite{rahaman2019spectral,xu2020frequency}. The last three factors apply to any attention-based surrogate trained with a mean-squared objective, so our measurements establish consistency with this account rather than isolating the bottleneck's contribution.

If the bottleneck were the sole cause, widening it should recover the missing scales. Sweeping the number of latents from 8 to 512 with all other settings fixed leaves every band almost flat (Fig.~\ref{fig:num-latents} in the appendix), consistent with the saturation reported for Perceiver IO, the Set Transformer, and low-rank spatial attention~\cite{lee2019set,jaegle2021perceiverio,yang2026simple}. Training does not exploit the capacity a wider bottleneck offers, as the loss- and optimization-side factors predict; the practical route to the high band is a second operator class rather than a wider bottleneck.

\subsection{Message Passing Fails from the Other End}
\label{sec:mgn-analysis}

The MGN column of Fig.~\ref{fig:training-dynamics} shows a different failure. The error contracts quickly over the first 150 epochs, with the larger early gains toward the middle and higher wavenumbers, and then plateaus; from epoch 300 onward the curves are nearly indistinguishable, and no band, including the lowest, improves further. Where \rwp{}'s low band keeps contracting throughout training, MGN stalls everywhere, at an error several times higher across the spectrum.

This is the failure the message-passing literature predicts for a steady problem. A message-passing layer is a local stencil that resolves structure near the mesh scale cheaply, which is consistent with the early gains sitting in the higher bands, but information travels only one edge per layer, and on steady and elliptic problems, where boundary data shapes the interior globally, the number of sweeps must grow with the ratio of domain size to mesh spacing before solutions stop degrading~\cite{brandstetter2022message}. On our meshes the graph diameter far exceeds the depth of a typical MGN processor, so the domain-scale content is out of reach; nor is deepening the stack a clean fix, since squeezing an exponentially growing neighborhood through fixed-width messages degrades the information that does arrive. Since the nonlinear couplings of the flow tie the wavenumber bands together, the stalled low band plausibly holds back the rest of the spectrum, though our runs do not isolate this mechanism.

The two pure processors therefore fail from opposite ends, one shedding the scales it cannot afford, the other starving the scales it cannot reach, and the remedy each needs is what the other provides.

\subsection{Combining Local and Global Processing}
\label{sec:rwr-analysis}

\begin{figure}[htbp]
\begin{minipage}[c]{0.50\linewidth}
\centering
\footnotesize
\setlength{\tabcolsep}{2.5pt}
\begin{tabular}{@{}lcccc@{}}
\toprule
& Transolver & GeoTransolver & \rwp{} & \rwr{} \\
\midrule
$u_x$ & 0.078\,[1.01] & 0.098\,[1.28] & 0.090\,[1.17] & \textbf{0.077} \\
$u_x$ grad. & 0.174\,[1.18] & 0.226\,[1.54] & 0.168\,[1.14] & \textbf{0.147} \\
$u_z$ & 0.080\,[1.11] & 0.104\,[1.45] & 0.091\,[1.27] & \textbf{0.072} \\
$u_z$ grad. & 0.207\,[1.27] & 0.286\,[1.75] & 0.218\,[1.33] & \textbf{0.164} \\
\addlinespace[2pt]
$\omega_y$ & 0.143\,[1.23] & 0.212\,[1.82] & 0.138\,[1.19] & \textbf{0.117} \\
$\tau_{w,x}$ & 0.170\,[1.30] & 0.246\,[1.88] & 0.157\,[1.20] & \textbf{0.131} \\
$\tau_{w,z}$ & 0.231\,[1.42] & 0.363\,[2.23] & 0.247\,[1.51] & \textbf{0.163} \\
\bottomrule
\end{tabular}
\captionof{table}{Held-out test errors on the branched pipe at matched training budget: relative $L_2$ error of each velocity component and of its spatial gradients, and relative $L_2$ error of the derived quantities, the in-plane vorticity $\omega_y$ and the wall shear stress components $\tau_{w,i} = \mu\,\partial u_i/\partial n$ evaluated at wall boundary points. Bracketed values give the ratio of each model's error to \rwr{}'s. Metric definitions are given in App.~\ref{app:metrics}.}
\label{tab:endpoint}
\end{minipage}\hfill
\begin{minipage}[c]{0.47\linewidth}
\centering
\includegraphics[width=\linewidth]{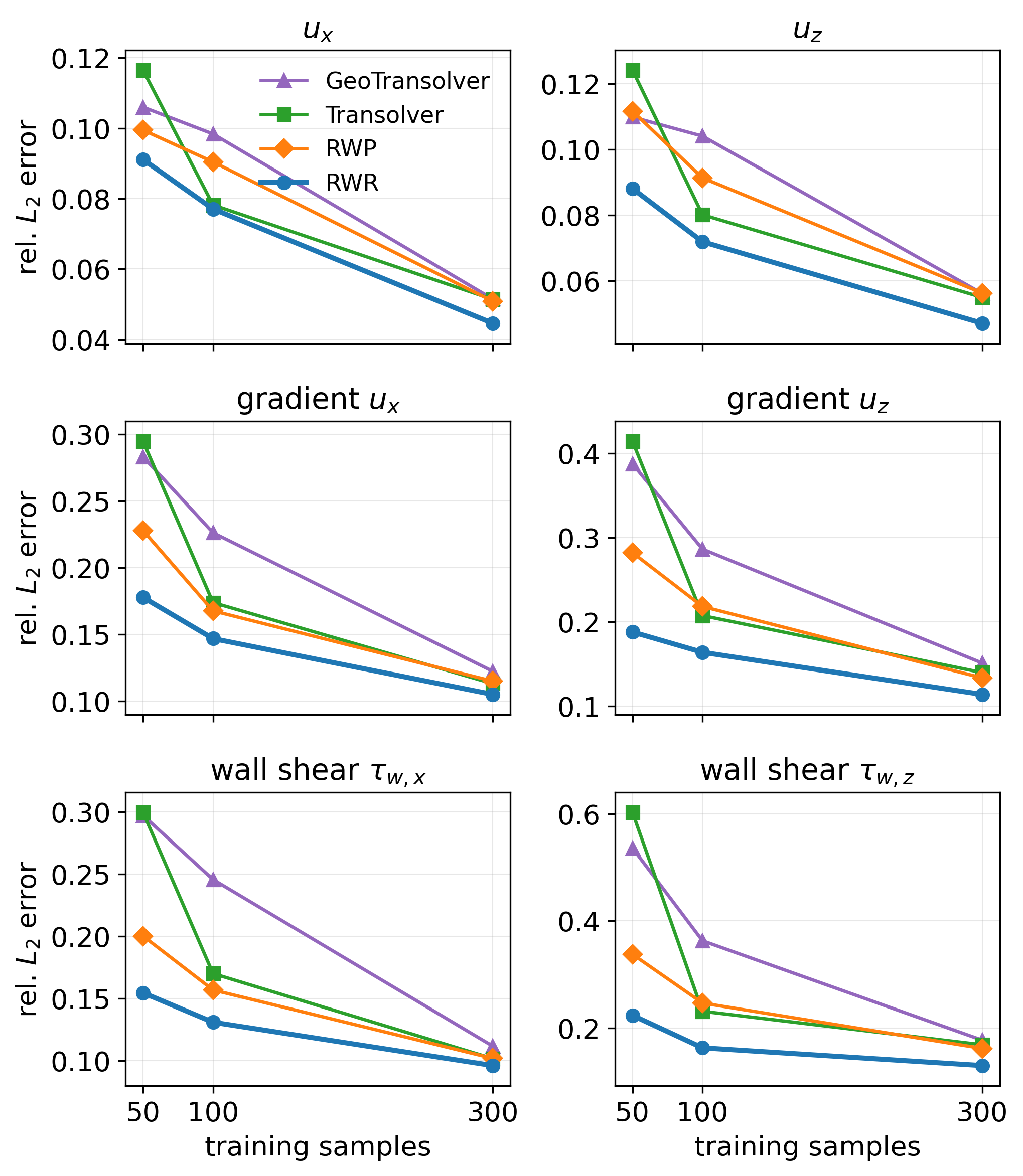}
\caption{Held-out test error on the branched pipe as a function of the training-set size for the bulk relative $L_2$ (top row), the gradient relative $L_2$ (middle row), and the wall shear stress (bottom row) of the two velocity components.}
\label{fig:data-efficiency}
\end{minipage}
\end{figure}

The \rwr{} column of Fig.~\ref{fig:training-dynamics} completes the picture. The interleaved processor contracts the error across the entire spectrum throughout training, with neither the plateau of MGN nor the high-band stagnation of \rwp{}. Table~\ref{tab:endpoint} compares the fully trained \rwr{} with the three latent-attention baselines at matched training budget, and the errors split along the expected lines. The upper rows report each velocity component and its spatial gradients, the lower rows the derived vorticity and wall shear stress, and the bracketed factors give each baseline's error relative to \rwr{}.

Moving from the bulk errors to the gradients and on to the derived wall quantities, the latent-attention models fall progressively further behind \rwr{}, and Fig.~\ref{fig:training-dynamics} shows the same progression spectrally, with the deficit growing toward the high wavenumbers. In the opposite direction the ordering softens, since the strongest latent-attention baselines still match \rwr{} at the lowest wavenumbers (Table~\ref{tab:full-metrics} in the appendix). Both halves are consistent with the preceding subsections. The latent-attention path remains the right tool for the smooth content, relaxation is what buys the fine scales, and the interleaved processor is the only one that holds both ends at once. The gradient and wall quantities deserve the emphasis, because they are the engineering quantities defined in the introduction, and bulk errors are documented to be poor proxies for them. It is precisely here that the combination delivers its largest margins, and the pointwise diagnostics (Figs.~\ref{fig:wall-parity} and \ref{fig:wall-distance} in the appendix) read the same, with \rwr{} attaining the highest wall-normal derivative parity of the four models and the lowest near-wall error on the velocity components.

The results so far establish that both operator classes are needed, but not that they must alternate, since a stacked composition that runs all relaxation before or after the latent-attention iterations contains the same two mechanisms. Comparing four allocations of an identical budget of eight relaxation sweeps and four latent-attention iterations, the two interleaved configurations improve on the two stacked orderings on every field, though all four land within a narrow range (Table~\ref{tab:interleaving} in the appendix). Once both operator classes are present at a sensible budget, the composition refines the picture rather than redrawing it.

\subsection{\rwr{} Is Data-Efficient}
\label{sec:data-efficiency}

The comparisons so far hold the training set fixed at one hundred samples. Fig.~\ref{fig:data-efficiency} varies the budget from 50 to 300 and tracks the bulk, gradient, and wall shear errors of the four models. \rwr{} attains the lowest error on every quantity at every budget, and its curves are the flattest, especially for the engineering quantities, so the margins are widest where data is scarcest. Trained on 50 samples, \rwr{} already matches or improves on every latent-attention baseline trained on twice as much data on the wall shear stress components. Since a training sample here is a full solver run, the scarce side of the sweep is where these surrogates actually operate, and it is there that \rwr{} pays the most.

\subsection{Benchmarks}
\label{sec:benchmarks}

We train Transolver, GeoTransolver, \rwp{}, and \rwr{} on the branched pipe and the two centrifugal pump datasets, each at three training-set sizes. Table~\ref{tab:internal} reports the mean test Normalized MAE (NMAE) over the output fields; per-field values, parameter counts, memory, and training cost are given in Table~\ref{tab:internal-full} in the appendix. The ordering of Fig.~\ref{fig:data-efficiency} on the branched pipe carries over to the industry-scale meshes, \rwr{} attains the lowest mean error in every cell, and the margins widen toward the smallest budgets, extending the data-efficiency picture of Sec.~\ref{sec:data-efficiency} across mesh scales. Qualitative predictions with pointwise error maps on held-out test cases of all three datasets are shown in Figs.~\ref{fig:pred-pipe-u}--\ref{fig:pred-full-p} in the appendix.

\begin{table}[htbp]
\footnotesize
\setlength{\tabcolsep}{2pt}
\begin{minipage}[t]{0.505\linewidth}
\centering
\resizebox{\linewidth}{!}{%
\begin{tabular}{@{}lccc@{}}
\toprule
& Branched pipe & Pump casing & Pump full \\
Model & (60k) & (270k) & (1.4M) \\
\midrule
Transolver & 11.3\,$|$\,7.5\,$|$\,4.8 & 15.6\,$|$\,10.9\,$|$\,7.9 & 37.1\,$|$\,23.2\,$|$\,15.9 \\
GeoTransolver & 10.5\,$|$\,8.3\,$|$\,5.5 & 12.1\,$|$\,\phantom{0}9.3\,$|$\,6.9 & 32.2\,$|$\,21.1\,$|$\,14.1 \\
\rwp{} & 10.3\,$|$\,7.3\,$|$\,5.0 & 11.4\,$|$\,\phantom{0}8.1\,$|$\,6.3 & 21.8\,$|$\,17.3\,$|$\,14.4 \\
\rwr{} & \phantom{0}\textbf{8.4}\,$|$\,\textbf{6.9}\,$|$\,\textbf{4.7} & \textbf{10.8}\,$|$\,\phantom{0}\textbf{7.7}\,$|$\,\textbf{5.4} & \textbf{19.3}\,$|$\,\textbf{15.2}\,$|$\,\textbf{12.7} \\
\bottomrule
\end{tabular}}
\caption{Benchmarks on the branched pipe and pump datasets. Each cell reports the mean test NMAE ($\times 10^{-2}$) over the output fields at 50 $|$ 100 $|$ 300 training samples, for models trained in our pipeline; model and training details are given in App.~\ref{app:internal-benchmark}. Bold marks the best value per dataset and budget.}
\label{tab:internal}
\end{minipage}\hfill
\begin{minipage}[t]{0.47\linewidth}
\centering
\resizebox{\linewidth}{!}{%
\begin{tabular}{@{}lccccc@{}}
\toprule
& & & \multicolumn{3}{c}{Geo-FNO} \\
\cmidrule(lr){4-6}
Model & AirfRANS & AhmedML & airfoil & pipe & Darcy \\
\midrule
MGN & 141 & 122 & 60.4 & 205 & 98.7 \\
MeshTransolver & 19.6 & 9.8 & 2.32 & \textbf{0.403} & 0.678 \\
Transolver & 21.4 & 9.3 & 1.99 & 0.643 & 0.356 \\
GeoTransolver & 20.9 & 8.6 & 2.15 & 0.845 & 0.412 \\
\rwp{} & 29.9 & 10.8 & 1.78 & 0.554 & 0.218 \\
\rwr{} & \textbf{15.7} & \textbf{7.2} & \textbf{1.58} & 0.409 & \textbf{0.190} \\
\bottomrule
\end{tabular}}
\caption{Best validation MSE for additional benchmarks ($\times 10^{-3}$) on normalized fields. Bold marks the best value per dataset.}
\label{tab:external}
\end{minipage}
\end{table}

We trained models additionally on AirfRANS~\cite{bonnet2022airfrans}, the surface mesh component of the AhmedML~\cite{ashton2024ahmedml} dataset, and the airfoil, pipe, and Darcy problems of the Geo-FNO suite~\cite{li2023geofno,li2021fourier}. These datasets provide a combination of irregular/regular 2D meshes and a 3D surface mesh to complement the 3D meshes of Table~\ref{tab:internal}. Table~\ref{tab:external} reports the best validation Mean Squared Error (MSE) on normalized fields for MGN, MeshTransolver, Transolver, GeoTransolver, \rwp{}, and \rwr{}; datasets, model configurations, and the full protocols are given in App.~\ref{app:public}.

The two tables read consistently. Since these benchmarks use a single random seed, we make only general claims about the comparisons. \rwr{} is the best-performing model on all datasets of Table~\ref{tab:internal} and on all additional benchmarks except the Geo-FNO pipe, where MeshTransolver is marginally better. As discussed in Sec.~\ref{sec:rwr-analysis} and demonstrated in Fig.~\ref{fig:training-dynamics}, we expect both \rwp{} and MGN to perform worse than \rwr{}. Indeed, MGN is consistently and significantly the worst-performing model on all benchmarks. \rwp{} performs below \rwr{} in all cases, yet remains in the performance range of Transolver and GeoTransolver, exceeding both on the three Geo-FNO problems. Finally, we see that the MGN-Transolver hybrid structure of MeshTransolver offers competitive performance on these additional benchmarks.

Importantly, these comparisons are on bulk metrics alone. Table~\ref{tab:endpoint} showed that where \rwr{} improves the bulk error over a pure latent-attention model, its gains on the gradient and wall metrics are larger still, so we expect the margins reported here to widen further on those engineering quantities (Sec.~\ref{sec:rwr-analysis}). Future work will demonstrate detailed comparisons of additional engineering quantities on industrial-scale datasets for further applications using the \rwr{} architecture.
% ─────────────────────────────────────────────────────────────────────────────
\section{Conclusion}
\label{sec:conclusion}

Global latent-token surrogates and local message passing fail on complementary ends of the error spectrum. We placed the recent global models in a single bottlenecked self-attention template and showed that on an industrial steady-flow benchmark they learn overwhelmingly at large spatial scales, a behavior consistent with several compounding causes, and one that widening the bottleneck does not repair. Message passing fails in the opposite way, gaining early on the fine scales but stalling across the spectrum when its receptive field cannot cover the domain. \rwr{} interleaves the two mechanisms as multigrid alternates coarse correction and relaxation, exposes the mix as a single configuration $(m_1, r, m_2)$ with MGN and \rwp{} as special cases, and contracts the error across the entire spectrum, with its largest margins on the engineering quantities and at the smallest training budgets, where CAE surrogates actually operate. On public benchmarks it is the most accurate model in nearly every comparison, and on industry-scale meshes it attains the lowest error at every training budget.

\paragraph{Limitations} Our analysis establishes consistency rather than causal isolation, since the compounding factors behind the low-pass behavior are not disentangled by our experiments, and narrowing them down requires controlled interventions on the loss, the attention, and the bottleneck individually. The behavior also attenuates as the training set grows, so part of it belongs to the data regime rather than to the architecture alone, and separating the two requires sweeps beyond the budgets we report. The spectral analysis rests on a single steady RANS benchmark, and our datasets are steady throughout, so transient rollouts stress failure modes we do not measure. The multigrid reading is an analogy of roles rather than a convergence theorem. Finally, our comparisons with hybrid models that combine message passing with attention are limited to our own implementation of the MeshTransolver-style sandwich, whose reference code is not public, and a broader comparison with this class remains open.

% ─────────────────────────────────────────────────────────────────────────────

% ─────────────────────────────────────────────────────────────────────────────
\paragraph{Acknowledgements} The authors acknowledge the contributions of Diego Andrade, Jau-Uei Chen, and Oliver Littlewood for industrially-motivated simulation dataset preparation.

\PLrefheading
\bibliography{refs}

@inproceedings{jaegle2021perceiver,
  title={Perceiver: General Perception with Iterative Attention},
  author={Jaegle, Andrew and Gimeno, Felix and Brock, Andrew and Zisserman, Andrew and Vinyals, Oriol and Carreira, Joao},
  booktitle={Proceedings of the 38th International Conference on Machine Learning},
  series={PMLR},
  year={2021}
}

@inproceedings{jaegle2021perceiverio,
  title={Perceiver {IO}: A General Architecture for Structured Inputs and Outputs},
  author={Jaegle, Andrew and Borgeaud, Sebastian and Alayrac, Jean-Baptiste and Doersch, Carl and Ionescu, Catalin and Ding, David and Koppula, Skanda and Zoran, Daniel and Brock, Andrew and Shelhamer, Evan and others},
  booktitle={International Conference on Learning Representations},
  year={2022}
}

@inproceedings{lee2019set,
  title={Set Transformer: A Framework for Attention-Based Permutation-Invariant Neural Networks},
  author={Lee, Juho and Lee, Yoonho and Kim, Jungtaek and Kosiorek, Adam R. and Choi, Seungjin and Teh, Yee Whye},
  booktitle={Proceedings of the 36th International Conference on Machine Learning},
  series={PMLR},
  year={2019}
}

@inproceedings{jabri2023rin,
  title={Scalable Adaptive Computation for Iterative Generation},
  author={Jabri, Allan and Fleet, David J. and Chen, Ting},
  booktitle={Proceedings of the 40th International Conference on Machine Learning},
  series={PMLR},
  year={2023}
}

@inproceedings{wu2024transolver,
  title={Transolver: A Fast Transformer Solver for {PDE}s on General Geometries},
  author={Wu, Haixu and Luo, Huakun and Wang, Haowen and Wang, Jianmin and Long, Mingsheng},
  booktitle={Proceedings of the 41st International Conference on Machine Learning},
  series={PMLR},
  year={2024}
}

@inproceedings{luo2025transolverpp,
  title={Transolver++: An Accurate Neural Solver for {PDE}s on Million-Scale Geometries},
  author={Luo, Huakun and Wu, Haixu and Zhou, Hang and Xing, Lanxiang and Di, Yichen and Wang, Jianmin and Long, Mingsheng},
  booktitle={Proceedings of the 42nd International Conference on Machine Learning},
  series={PMLR},
  year={2025}
}

@misc{puri2025flare,
  title={{FLARE}: Fast Low-Rank Attention Routing Engine},
  author={Puri, Vedant and Joglekar, Aditya and Bandreddi, Sri Datta Ganesh and Ferguson, Kevin and Chen, Yu-hsuan and Zhang, Yongjie Jessica and Kara, Levent Burak},
  year={2025},
  eprint={2508.12594},
  archivePrefix={arXiv}
}

@misc{adams2025geotransolver,
  title={{GeoTransolver}: Learning Physics on Irregular Domains Using Multi-Scale Geometry Aware Physics Attention Transformer},
  author={Adams, Corey and Ranade, Rishikesh and Nidhan, Sheel and Nabian, Mohammad Amin and Akhare, Deepak and Cherukuri, Ram and Choudhry, Sanjay},
  year={2025},
  eprint={2512.20399},
  archivePrefix={arXiv}
}

@misc{yang2026simple,
  title={Simple yet Effective: Low-Rank Spatial Attention for Neural Operators},
  author={Yang, Zherui and Xin, Haiyang and Du, Tao and Liu, Ligang},
  year={2026},
  eprint={2604.03582},
  archivePrefix={arXiv}
}

@inproceedings{wang2022antioversmoothing,
  title={Anti-Oversmoothing in Deep Vision Transformers via the {Fourier} Domain Analysis: From Theory to Practice},
  author={Wang, Peihao and Zheng, Wenqing and Chen, Tianlong and Wang, Zhangyang},
  booktitle={International Conference on Learning Representations},
  year={2022}
}

@inproceedings{park2022how,
  title={How Do Vision Transformers Work?},
  author={Park, Namuk and Kim, Songkuk},
  booktitle={International Conference on Learning Representations},
  year={2022}
}

@inproceedings{tsai2019transformer,
  title={Transformer Dissection: A Unified Understanding of Transformer's Attention via the Lens of Kernel},
  author={Tsai, Yao-Hung Hubert and Bai, Shaojie and Yamada, Makoto and Morency, Louis-Philippe and Salakhutdinov, Ruslan},
  booktitle={Proceedings of the 2019 Conference on Empirical Methods in Natural Language Processing},
  year={2019}
}

@inproceedings{rahaman2019spectral,
  title={On the Spectral Bias of Neural Networks},
  author={Rahaman, Nasim and Baratin, Aristide and Arpit, Devansh and Draxler, Felix and Lin, Min and Hamprecht, Fred A. and Bengio, Yoshua and Courville, Aaron},
  booktitle={Proceedings of the 36th International Conference on Machine Learning},
  series={PMLR},
  year={2019}
}

@article{xu2020frequency,
  title={Frequency Principle: {Fourier} Analysis Sheds Light on Deep Neural Networks},
  author={Xu, Zhi-Qin John and Zhang, Yaoyu and Luo, Tao and Xiao, Yanyang and Ma, Zheng},
  journal={Communications in Computational Physics},
  volume={28},
  number={5},
  pages={1746--1767},
  year={2020}
}

@inproceedings{pfaff2020mgn,
  title={Learning Mesh-Based Simulation with Graph Networks},
  author={Pfaff, Tobias and Fortunato, Meire and Sanchez-Gonzalez, Alvaro and Battaglia, Peter},
  booktitle={International Conference on Learning Representations},
  year={2021}
}

@inproceedings{fortunato2022multiscale,
  title={MultiScale MeshGraphNets},
  author={Fortunato, Meire and Pfaff, Tobias and Wirnsberger, Peter and Pritzel, Alexander and Battaglia, Peter},
  booktitle={2nd AI for Science Workshop at the 39th International Conference on Machine Learning},
  year={2022}
}

@inproceedings{cao2023efficient,
  title={Efficient Learning of Mesh-Based Physical Simulation with Bi-Stride Multi-Scale Graph Neural Network},
  author={Cao, Yadi and Chai, Menglei and Li, Minchen and Jiang, Chenfanfu},
  booktitle={Proceedings of the 40th International Conference on Machine Learning},
  series={PMLR},
  year={2023}
}

@inproceedings{brandstetter2022message,
  title={Message Passing Neural {PDE} Solvers},
  author={Brandstetter, Johannes and Worrall, Daniel E. and Welling, Max},
  booktitle={International Conference on Learning Representations},
  year={2022}
}

@inproceedings{alon2021on,
  title={On the Bottleneck of Graph Neural Networks and Its Practical Implications},
  author={Alon, Uri and Yahav, Eran},
  booktitle={International Conference on Learning Representations},
  year={2021}
}

@inproceedings{topping2022understanding,
  title={Understanding Over-Squashing and Bottlenecks on Graphs via Curvature},
  author={Topping, Jake and Di Giovanni, Francesco and Chamberlain, Benjamin P. and Dong, Xiaowen and Bronstein, Michael M.},
  booktitle={International Conference on Learning Representations},
  year={2022}
}

@inproceedings{lippe2023pderefiner,
  title={{PDE}-Refiner: Achieving Accurate Long Rollouts with Neural {PDE} Solvers},
  author={Lippe, Phillip and Veeling, Bastiaan S. and Perdikaris, Paris and Turner, Richard E. and Brandstetter, Johannes},
  booktitle={Advances in Neural Information Processing Systems},
  year={2023}
}

@inproceedings{li2023gino,
  title={Geometry-Informed Neural Operator for Large-Scale {3D} {PDE}s},
  author={Li, Zongyi and Kovachki, Nikola Borislavov and Choy, Chris and Li, Boyi and Kossaifi, Jean and Otta, Shourya Prakash and Nabian, Mohammad Amin and Stadler, Maximilian and Hundt, Christian and Azizzadenesheli, Kamyar and Anandkumar, Anima},
  booktitle={Advances in Neural Information Processing Systems},
  year={2023}
}

@inproceedings{alkin2024upt,
  title={Universal Physics Transformers: A Framework for Efficiently Scaling Neural Operators},
  author={Alkin, Benedikt and F{\"u}rst, Andreas and Schmid, Simon and Gruber, Lukas and Holzleitner, Markus and Brandstetter, Johannes},
  booktitle={Advances in Neural Information Processing Systems},
  year={2024}
}

@misc{wen2025gaot,
  title={Geometry Aware Operator Transformer as an Efficient and Accurate Neural Surrogate for {PDE}s on Arbitrary Domains},
  author={Wen, Shizheng and Mousavi, Sepehr and others},
  year={2025},
  eprint={2505.18781},
  archivePrefix={arXiv}
}

@book{kress2014linear,
  title={Linear Integral Equations},
  author={Kress, Rainer},
  edition={3rd},
  publisher={Springer},
  year={2014}
}

@misc{duthe2025graph,
  title={Graph Transformers for Inverse Physics: Reconstructing Flows Around Arbitrary {2D} Airfoils},
  author={Duth{\'e}, Gregory and others},
  year={2025},
  eprint={2501.17081},
  archivePrefix={arXiv}
}

@inproceedings{tancik2020rff,
  title={Fourier Features Let Networks Learn High Frequency Functions in Low Dimensional Domains},
  author={Tancik, Matthew and Srinivasan, Pratul P. and Mildenhall, Ben and Fridovich-Keil, Sara and Raghavan, Nithin and Singhal, Utkarsh and Ramamoorthi, Ravi and Barron, Jonathan T. and Ng, Ren},
  booktitle={Advances in Neural Information Processing Systems},
  year={2020}
}

@inproceedings{gilmer2017neural,
  title={Neural Message Passing for Quantum Chemistry},
  author={Gilmer, Justin and Schoenholz, Samuel S. and Riley, Patrick F. and Vinyals, Oriol and Dahl, George E.},
  booktitle={Proceedings of the 34th International Conference on Machine Learning},
  series={PMLR},
  year={2017}
}

@inproceedings{cai2023connection,
  title={On the Connection Between {MPNN} and Graph Transformer},
  author={Cai, Chen and Hy, Truong Son and Yu, Rose and Wang, Yusu},
  booktitle={Proceedings of the 40th International Conference on Machine Learning},
  series={PMLR},
  year={2023}
}

@inproceedings{wu2021representing,
  title={Representing Long-Range Context for Graph Neural Networks with Global Attention},
  author={Wu, Zhanghao and Jain, Paras and Wright, Matthew A. and Mirhoseini, Azalia and Gonzalez, Joseph E. and Stoica, Ion},
  booktitle={Advances in Neural Information Processing Systems},
  year={2021}
}

@inproceedings{rampasek2022recipe,
  title={Recipe for a General, Powerful, Scalable Graph Transformer},
  author={Ramp{\'a}{\v s}ek, Ladislav and Galkin, Mikhail and Dwivedi, Vijay Prakash and Luu, Anh Tuan and Wolf, Guy and Beaini, Dominique},
  booktitle={Advances in Neural Information Processing Systems},
  year={2022}
}

@inproceedings{shirzad2023exphormer,
  title={Exphormer: Sparse Transformers for Graphs},
  author={Shirzad, Hamed and Velingker, Ameya and Venkatachalam, Balaji and Sutherland, Danica J. and Sinop, Ali Kemal},
  booktitle={Proceedings of the 40th International Conference on Machine Learning},
  series={PMLR},
  year={2023}
}

@inproceedings{kreuzer2021rethinking,
  title={Rethinking Graph Transformers with Spectral Attention},
  author={Kreuzer, Devin and Beaini, Dominique and Hamilton, William L. and L{\'e}tourneau, Vincent and Tossou, Prudencio},
  booktitle={Advances in Neural Information Processing Systems},
  year={2021}
}

@inproceedings{ying2021transformers,
  title={Do Transformers Really Perform Badly for Graph Representation?},
  author={Ying, Chengxuan and Cai, Tianle and Luo, Shengjie and Zheng, Shuxin and Ke, Guolin and He, Di and Shen, Yanming and Liu, Tie-Yan},
  booktitle={Advances in Neural Information Processing Systems},
  year={2021}
}

@inproceedings{janny2023eagle,
  title={{EAGLE}: Large-Scale Learning of Turbulent Fluid Dynamics with Mesh Transformers},
  author={Janny, Steeven and Beneteau, Aur{\'e}lien and Nadri, Madiha and Digne, Julie and Thome, Nicolas and Wolf, Christian},
  booktitle={International Conference on Learning Representations},
  year={2023}
}

@inproceedings{wang2024beno,
  title={{BENO}: Boundary-Embedded Neural Operators for Elliptic {PDE}s},
  author={Wang, Haixin and Li, Jiaxin and Dwivedi, Anubhav and Hara, Kentaro and Wu, Tailin},
  booktitle={International Conference on Learning Representations},
  year={2024}
}

@inproceedings{wang2024latent,
  title={Latent Neural Operator for Solving Forward and Inverse {PDE} Problems},
  author={Wang, Tian and Wang, Chuang},
  booktitle={Advances in Neural Information Processing Systems},
  year={2024}
}

@article{gladstone2024mesh,
  title={Mesh-Based {GNN} Surrogates for Time-Independent {PDE}s},
  author={Gladstone, Rini Jasmine and Rahmani, Helia and Suryakumar, Vignesh and Meidani, Hadi and D'Elia, Marta and Zareei, Ahmad},
  journal={Scientific Reports},
  volume={14},
  pages={3394},
  year={2024}
}

@inproceedings{ripken2023multiscale,
  title={Multiscale Neural Operators for Solving Time-Independent {PDE}s},
  author={Ripken, Winfried and Coiffard, Lisa and Pieper, Felix and Dziadzio, Sebastian},
  booktitle={The Symbiosis of Deep Learning and Differential Equations {III} Workshop at NeurIPS},
  year={2023}
}

@misc{ramezankhani2025gito,
  title={{GITO}: Graph-Informed Transformer Operator for Learning Complex Partial Differential Equations},
  author={Ramezankhani, Milad and Patel, Janak M. and Deodhar, Anirudh and Birru, Dagnachew},
  year={2025},
  eprint={2506.13906},
  archivePrefix={arXiv}
}

@misc{iparraguirre2026mgnt,
  title={MeshGraphNet-Transformer: Scalable Mesh-Based Learned Simulation for Solid Mechanics},
  author={Iparraguirre, Mat{\'\i}as M. and Alfaro, Ic{\'\i}ar and Gonzalez, David and Cueto, El{\'\i}as},
  year={2026},
  eprint={2601.23177},
  archivePrefix={arXiv}
}

@misc{curtosi2026crash,
  title={Crash Assessment via Mesh-Based Graph Neural Networks and Physics-Aware Attention},
  author={Curtosi, Giovanni and Ruiz Ruiz, Carlos M. and Cavaliere, Fabio and Larr{\'a}yoz Izcara, Xabier},
  year={2026},
  eprint={2605.11784},
  archivePrefix={arXiv}
}

@inproceedings{takamoto2022pdebench,
  title={{PDEBench}: An Extensive Benchmark for Scientific Machine Learning},
  author={Takamoto, Makoto and Praditia, Timothy and Leiteritz, Raphael and MacKinlay, Dan and Alesiani, Francesco and Pfl{\"u}ger, Dirk and Niepert, Mathias},
  booktitle={Advances in Neural Information Processing Systems},
  year={2022}
}

@inproceedings{serrano2024aroma,
  title={{AROMA}: Preserving Spatial Structure for Latent {PDE} Modeling with Local Neural Fields},
  author={Serrano, Louis and Wang, Thomas X. and Le Naour, Etienne and Vittaut, Jean-No{\"e}l and Gallinari, Patrick},
  booktitle={Advances in Neural Information Processing Systems},
  year={2024}
}

@misc{qin2024toward,
  title={Toward a Better Understanding of {Fourier} Neural Operators from a Spectral Perspective},
  author={Qin, Shaoxiang and Lyu, Fuyuan and Peng, Wenhui and Geng, Dingyang and Wang, Ju and Tang, Xing and Leroyer, Sylvie and Gao, Naiping and Liu, Xue and Wang, Liangzhu},
  year={2024},
  eprint={2404.07200},
  archivePrefix={arXiv}
}

@misc{khodakarami2025mitigating,
  title={Mitigating Spectral Bias in Neural Operators via High-Frequency Scaling for Physical Systems},
  author={Khodakarami, Siavash and Oommen, Vivek and Bora, Aniruddha and Karniadakis, George Em},
  year={2025},
  eprint={2503.13695},
  archivePrefix={arXiv}
}

@book{pope2000turbulent,
  title={Turbulent Flows},
  author={Pope, Stephen B.},
  publisher={Cambridge University Press},
  year={2000}
}

@article{batchelor1951pressure,
  title={Pressure Fluctuations in Isotropic Turbulence},
  author={Batchelor, George K.},
  journal={Mathematical Proceedings of the Cambridge Philosophical Society},
  volume={47},
  number={2},
  pages={359--374},
  year={1951}
}

@article{gotoh2001pressure,
  title={Pressure Spectrum in Homogeneous Turbulence},
  author={Gotoh, Toshiyuki and Fukayama, Daigen},
  journal={Physical Review Letters},
  volume={86},
  number={17},
  pages={3775--3778},
  year={2001}
}

@article{tsuji2003similarity,
  title={Similarity Scaling of Pressure Fluctuation in Turbulence},
  author={Tsuji, Yoshiyuki and Ishihara, Takashi},
  journal={Physical Review E},
  volume={68},
  pages={026309},
  year={2003}
}

@article{vlaykov2019pressure,
  title={On the Small-Scale Structure of Turbulence and Its Impact on the Pressure Field},
  author={Vlaykov, Dimitar G. and Wilczek, Michael},
  journal={Journal of Fluid Mechanics},
  volume={861},
  pages={422--446},
  year={2019}
}

@inproceedings{bonnet2022airfrans,
  title={{AirfRANS}: High Fidelity Computational Fluid Dynamics Dataset for Approximating {Reynolds-Averaged Navier--Stokes} Solutions},
  author={Bonnet, Florent and Mazari, Jocelyn Ahmed and Cinnella, Paola and Gallinari, Patrick},
  booktitle={Advances in Neural Information Processing Systems},
  year={2022}
}

@article{li2023geofno,
  title={Fourier Neural Operator with Learned Deformations for {PDE}s on General Geometries},
  author={Li, Zongyi and Huang, Daniel Zhengyu and Liu, Burigede and Anandkumar, Anima},
  journal={Journal of Machine Learning Research},
  volume={24},
  number={388},
  pages={1--26},
  year={2023}
}

@misc{ashton2024ahmedml,
  title={{AhmedML}: High-Fidelity Computational Fluid Dynamics Dataset for Incompressible, Low-Speed Bluff Body Aerodynamics},
  author={Ashton, Neil and Maddix, Danielle C. and Gundry, Samuel and Shabestari, Parisa M.},
  year={2024},
  eprint={2407.20801},
  archivePrefix={arXiv}
}

@misc{ashton2024drivaerml,
  title={{DrivAerML}: High-Fidelity Computational Fluid Dynamics Dataset for Road-Car External Aerodynamics},
  author={Ashton, Neil and Mockett, Charles and Fuchs, Marian and Fliessbach, Louis and Hetmann, Hendrik and Knacke, Thilo and Schonwald, Norbert and Skaperdas, Vangelis and Fotiadis, Grigoris and Walle, Astrid and Hupertz, Burkhard and Maddix, Danielle},
  year={2024},
  eprint={2408.11969},
  archivePrefix={arXiv}
}

@article{brandt1977multi,
  title={Multi-Level Adaptive Solutions to Boundary-Value Problems},
  author={Brandt, Achi},
  journal={Mathematics of Computation},
  volume={31},
  number={138},
  pages={333--390},
  year={1977}
}

@book{briggs2000multigrid,
  title={A Multigrid Tutorial},
  author={Briggs, William L. and Henson, Van Emden and McCormick, Steve F.},
  edition={2nd},
  publisher={SIAM},
  year={2000}
}

@inproceedings{li2021fourier,
  title={Fourier Neural Operator for Parametric Partial Differential Equations},
  author={Li, Zongyi and Kovachki, Nikola and Azizzadenesheli, Kamyar and Liu, Burigede and Bhattacharya, Kaushik and Stuart, Andrew and Anandkumar, Anima},
  booktitle={International Conference on Learning Representations},
  year={2021}
}

\clearpage
\appendix

\section{Datasets and Simulations}
\label{app:datasets}

Both environments are internal-flow cases solved with the Finite Volume Method (FVM), using a RANS turbulence model, a velocity-inlet and pressure-outlet boundary-condition pair, and no-slip walls. The branched pipe applies a one-seventh power-law velocity profile at the inlet; the pump applies a uniform inlet velocity and models the impeller rotation with the Moving Reference Frame (MRF) method. Both are steady-state simulations solved with an adaptive time-stepping strategy. Table~\ref{tab:sim-config} summarizes the simulation configurations, Table~\ref{tab:design-params} the design parameters sampled to generate each dataset, and Table~\ref{tab:dataset-spec} the modeling task and dataset specification shared by all models.

\begin{table}[htbp]
\centering
\small
\setlength{\tabcolsep}{3pt}
\begin{tabular}{@{}lccc@{}}
\toprule
& Branched pipe & \multicolumn{2}{c}{Centrifugal pump} \\
& & (casing) & (full) \\
\midrule
cell number & ${\sim}60$k & ${\sim}270$k & ${\sim}1.4$M \\
fluid & air & water & water \\
turbulence model & RANS & RANS & RANS \\
boundary conditions & \multicolumn{3}{c}{velocity inlet, pressure outlet, no-slip wall} \\
solver & \multicolumn{3}{c}{FVM (Fluent)} \\
special treatment & --- & MRF & MRF \\
\bottomrule
\end{tabular}
\caption{Summary of simulation configurations.}
\label{tab:sim-config}
\end{table}

\begin{table}[htbp]
\centering
\footnotesize
\setlength{\tabcolsep}{3pt}
\begin{tabular}{@{}lc@{}}
\toprule
Parameter & Range \\
\midrule
\multicolumn{2}{@{}l}{\emph{Branched pipe}} \\
angle of middle baffle plate & 0--60 (deg) \\
aperture length of inlet baffle plate & 20--37 (mm) \\
angle between ducts & 75--120 (deg) \\
curvature angle of duct branch & 90--180 (deg) \\
inlet mean velocity & 3--10 (m/s) \\
\addlinespace[2pt]
\multicolumn{2}{@{}l}{\emph{Centrifugal pump}} \\
inlet radius & 160--200 (mm) \\
impeller radius & 80--100 (mm) \\
blade pitch angle & 5--40 (deg) \\
inlet flow velocity & 1.5--3.5 (m/s) \\
\bottomrule
\end{tabular}
\caption{Design parameters used to generate the simulation datasets, with the range each parameter is sampled over.}
\label{tab:design-params}
\end{table}

\begin{table}[htbp]
\centering
\footnotesize
\setlength{\tabcolsep}{3pt}
\begin{tabular}{@{}lcc@{}}
\toprule
& Branched pipe & Centrifugal pump \\
\midrule
modeling task & steady-state & steady-state \\
total samples & 300 & 300 \\
training-set sizes & \multicolumn{2}{c}{50 / 100 / 300 samples} \\
train/val/test split & \multicolumn{2}{c}{0.7/0.15/0.15} \\
random seed & \multicolumn{2}{c}{42} \\
field normalization & \multicolumn{2}{c}{per-component standardization} \\
input features & \multicolumn{2}{c}{\makecell{coordinates, inlet/outlet\\ boundary conditions}} \\
input / processing domain & \multicolumn{2}{c}{full mesh} \\
modeled fields & $p$, $u_x$, $u_z$ & $u_x$, $u_y$, $u_z$, $p$ \\
\bottomrule
\end{tabular}
\caption{Modeling task and dataset specification, common to all models. The two pump extents share the same design of experiments and splits.}
\label{tab:dataset-spec}
\end{table}

\section{Metric Definitions}
\label{app:metrics}

All metrics compare a predicted field $u$ against the reference solution $u^*$ through the error $e = u - u^*$, in physical units after inverting the training normalization. Cell-centered data are converted to point data by volume-weighted averaging before all point-based metrics, with $R^2$ the one exception, and all means are unweighted. The analysis metrics are computed per case, field, and model on the held-out test split, capped to eight cases, and reported as means over these cases; the benchmark tables instead use the full 45-case test split. Truth and prediction always pass through the same operators, so operator error largely cancels in the relative metrics.

\paragraph{Relative $L_2$} $\sqrt{\operatorname{mean}(e^2)/\operatorname{mean}(u^{*2})}$ over all volume mesh points.

\paragraph{$R^2$} $1 - \operatorname{mean}(e^2)/\operatorname{var}(u^*)$, computed on the raw cell-centered arrays.

\paragraph{Gradients} Spatial gradients use a k-nearest-neighbor weighted least-squares operator rather than a mesh-based filter. For each point, the $k = 20$ nearest neighbors define displacements $d_j$ and value differences $\delta v_j$, and the gradient $g$ solves the ridge-regularized weighted normal equations $\bigl(\sum_j w_j d_j d_j^\top + \epsilon I\bigr)\, g = \sum_j w_j d_j\, \delta v_j$ with $w_j = 1/|d_j|$. The operator reproduces linear fields to machine precision, and the same operator is applied to truth and prediction. The gradient relative $L_2$ is $\sqrt{\operatorname{mean}|\nabla e|^2 / \operatorname{mean}|\nabla u^*|^2}$ with $|\cdot|$ the Euclidean norm.

\paragraph{Error spectrum} Spectral quantities live on a two-dimensional mid-plane slice. A $y$-normal plane at the domain's mid-$y$ is sampled on a uniform raster 512 pixels wide with square pixels, points outside the fluid are masked and zero-filled, and the two-dimensional FFT power is binned into integer radial wavenumber rings $k = \operatorname{round}\bigl(\sqrt{k_x^2 + k_z^2}\bigr)$, in cycles per domain extent. $E_{\mathrm{err}}(k)$ is the ring-summed power spectrum of the error field and $E_{\mathrm{true}}(k)$ that of the reference; wavenumbers above $k = 92$, where raster-interpolation noise floors the spectrum, are excluded, as is the DC ring.

\paragraph{Band-resolved fRMSE} fRMSE, the RMSE in Fourier space, is the band-relative error $\sqrt{\sum_{k \in \mathrm{band}} E_{\mathrm{err}}(k) / \sum_{k \in \mathrm{band}} E_{\mathrm{true}}(k)}$, 0 for a perfect prediction and 1 when the error is as energetic as the truth within the band. The cutoffs follow the PDEBench bins~\cite{takamoto2022pdebench}, $0 \le k \le 4$ (low), $5 \le k \le 12$ (mid), and $13 \le k \le 92$ (high).

\paragraph{Vorticity} $\omega_y = \partial u_x/\partial z - \partial u_z/\partial x$, computed with the same gradient operator and scored with the relative $L_2$ formula as a scalar field.

\paragraph{Wall shear stress} The reported components are the wall-normal derivative proxies $\tau_{w,i} = \mu\, (\nabla u_i \cdot \hat n)$ for $i \in \{x, z\}$, with $\mu$ the constant laminar viscosity and $\hat n$ the unit wall normals from the solver's face-area vectors; volume gradients are interpolated to each wall point as the mean over its four nearest volume points. This proxy is the dominant term of the wall shear stress for these components, and its truth values track the solver's stored wall shear. Scores are the relative $L_2$ and $R^2$ over all wall boundary points, and the parity plots of Fig.~\ref{fig:wall-parity} show predicted against true $\mu\, \partial u_i/\partial n$ pooled over cases.

\paragraph{Error versus wall distance} Wall distance $d_w$ is the solver's stored wall-distance field. Points fall into twelve quantile bins of $d_w$, once over the full range and once over the nearest-to-wall fifth of points, and each bin reports the mean absolute error, averaged over the test cases.

\paragraph{NMAE} Mean absolute error over all mesh cells, normalized by the standard deviation of the ground-truth field.

\paragraph{Training-dynamics normalization} The training-dynamics figures evaluate five evenly spaced checkpoints on four cases. Every curve is divided by \rwp{}'s first-checkpoint value, per field and wavenumber or band, so a value of 1 means as wrong as \rwp{} at initialization.

\section{Pressure and the Low-Pass Bias}
\label{app:pressure}
Across our experiments, pressure consistently behaves as the easiest field for the latent-attention models. In Table~\ref{tab:full-metrics} the strongest latent-attention baseline matches or exceeds \rwr{} on every pressure metric, and its band-resolved errors rise only mildly from the low to the high band on pressure while they steepen sharply on the velocity components. The same pattern has been reported for steady RANS airfoil flows, where surrogate models across architectures inferred surface pressure and lift coefficient far more accurately than wall shear stress and drag coefficient~\cite{bonnet2022airfrans}. The behavior has a physical reading. In incompressible flow, taking the divergence of the momentum equation yields a Poisson equation for the pressure, $\nabla^2 p = -\rho\, (\partial u_i/\partial x_j)(\partial u_j/\partial x_i)$, whose Green's-function solution expresses the pressure at a point as a domain-wide, $1/|x-y|$-weighted integral of the velocity-gradient source~\cite{pope2000turbulent}. The inverse Laplacian attenuates the high-wavenumber content of this source as $1/k^2$, and pressure is accordingly characterized as a long-range field~\cite{vlaykov2019pressure}. Consistently, in high-Reynolds-number turbulence the inertial-range pressure spectrum falls as $k^{-7/3}$, steeper than the $k^{-5/3}$ velocity spectrum~\cite{batchelor1951pressure,gotoh2001pressure}, although the exponent is only cleanly observed at high Reynolds number~\cite{tsuji2003similarity} and inertial-range scalings do not strictly apply to the steady RANS mean fields of our benchmark. The elliptic-smoothing mechanism, however, is Reynolds-independent, since the divergence of the steady RANS momentum equation yields the same Poisson structure for the mean pressure, with a source built from mean-velocity gradients and Reynolds-stress divergences. A model biased toward low wavenumbers therefore degrades pressure least and the velocity components most, whose spectra keep more of their content in the bands the bottleneck sheds.

\section{Additional Diagnostics}
\label{app:spectral}

\begin{figure*}[htbp]
\centering
\includegraphics[width=0.88\linewidth]{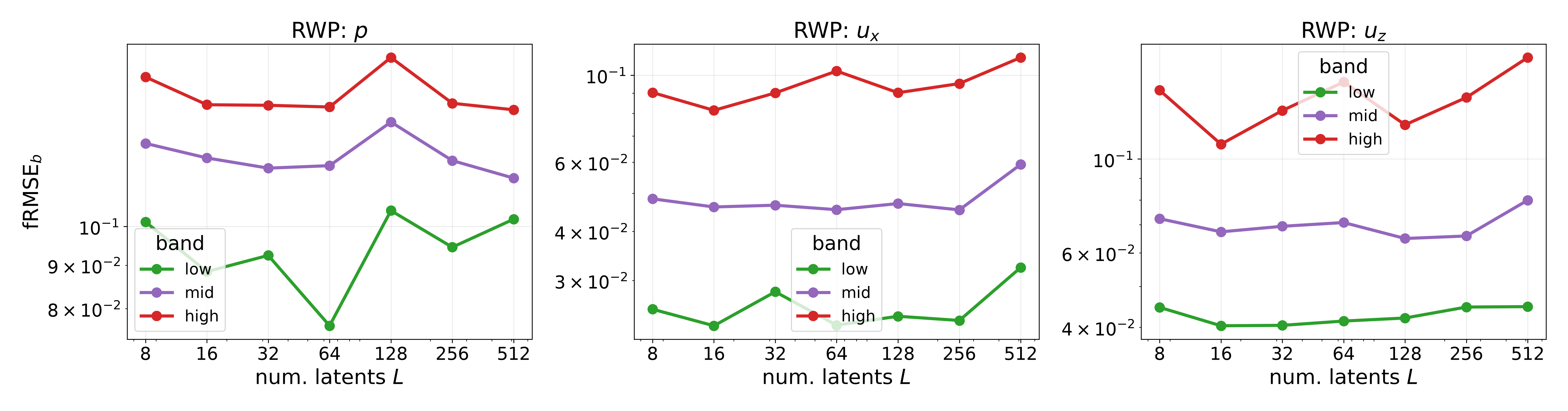}
\caption{Band-resolved test error of \rwp{} as a function of the number of latents $L$ on the branched pipe. Each panel shows the fRMSE over the low, mid, and high wavenumber bands for pressure and the two velocity components, with $L$ swept from 8 to 512 and all other settings fixed.}
\label{fig:num-latents}
\end{figure*}

\begin{figure*}[htbp]
\centering
\includegraphics[width=0.88\linewidth]{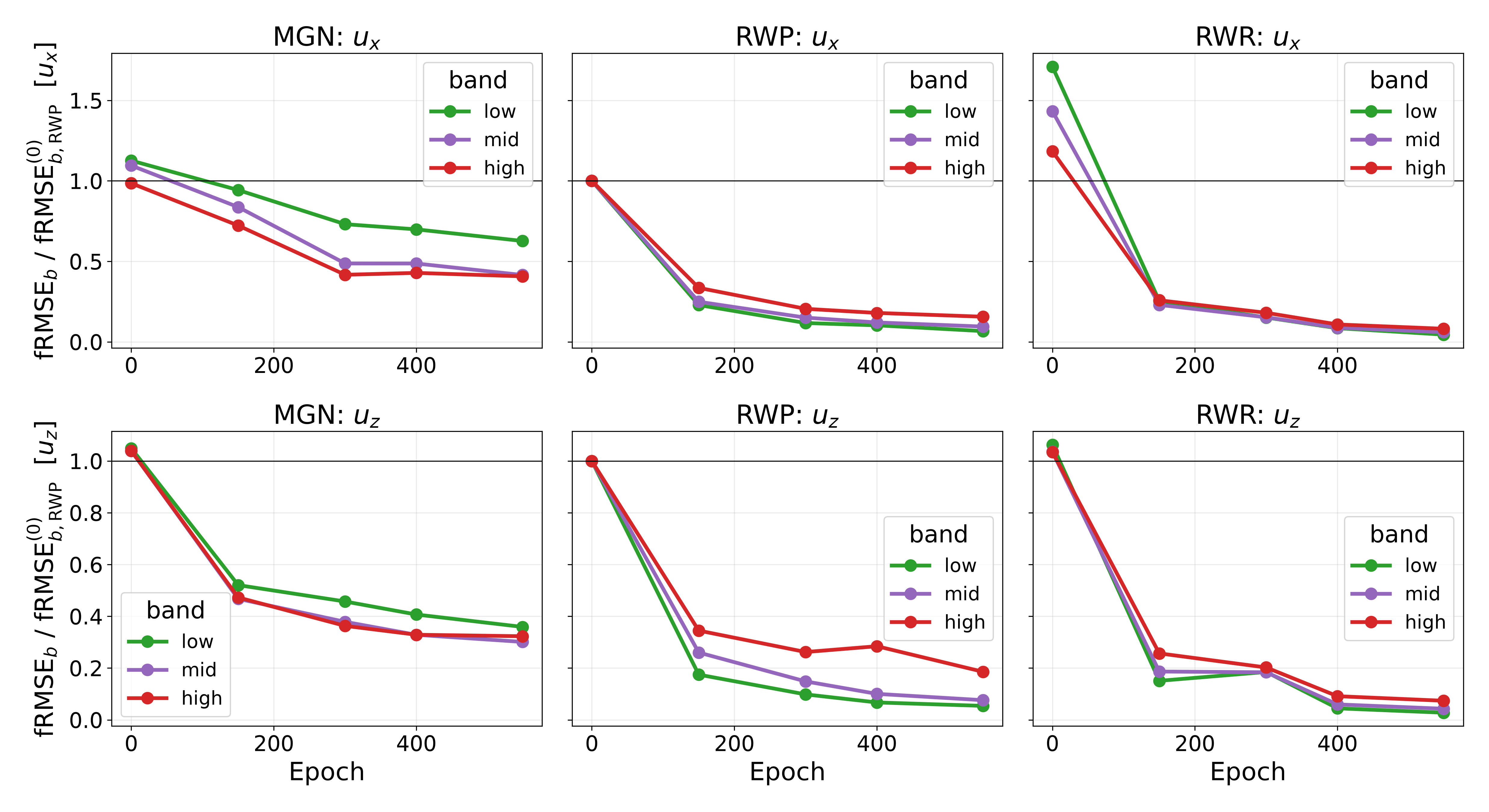}
\caption{Band-resolved training dynamics on the branched pipe. Each panel tracks the fRMSE of the low, mid, and high wavenumber bands over training for the two velocity components (rows) under MGN, \rwp{}, and \rwr{} (columns), normalized by the initial band value of \rwp{}.}
\label{fig:training-bands}
\end{figure*}

\subsection{Full Branched Pipe Metrics}
\label{app:full-metrics}

Table~\ref{tab:full-metrics} reports the complete per-field metrics behind Table~\ref{tab:endpoint}, the bulk and gradient relative $L_2$ errors together with the band-resolved fRMSE over the low, mid, and high wavenumber bands, for pressure and both velocity components.

\begin{table*}[htbp]
\centering
\small
\resizebox{\linewidth}{!}{%
\begin{tabular}{@{}llccccc@{}}
\toprule
& & \multicolumn{2}{c}{rel.\ $L_2$} & \multicolumn{3}{c}{fRMSE} \\
\cmidrule(lr){3-4} \cmidrule(lr){5-7}
Field & Model & field & grad. & low & mid & high \\
\midrule
$p$ & Transolver & \textbf{0.090}\,[0.85] & \textbf{0.165}\,[0.95] & \textbf{0.118}\,[0.99] & \textbf{0.129}\,[0.94] & \textbf{0.133}\,[0.87] \\
 & GeoTransolver & 0.135\,[1.28] & 0.212\,[1.22] & 0.159\,[1.33] & 0.168\,[1.22] & 0.174\,[1.14] \\
 & \rwp{} & 0.134\,[1.27] & 0.199\,[1.14] & 0.155\,[1.30] & 0.177\,[1.29] & 0.191\,[1.25] \\
 & \rwr{} & 0.106 & 0.174 & 0.119 & 0.137 & 0.153 \\
\addlinespace[2pt]
$u_x$ & Transolver & 0.078\,[1.01] & 0.174\,[1.18] & \textbf{0.035}\,[0.95] & 0.070\,[1.10] & 0.129\,[1.37] \\
 & GeoTransolver & 0.098\,[1.28] & 0.226\,[1.54] & 0.057\,[1.54] & 0.073\,[1.15] & 0.107\,[1.13] \\
 & \rwp{} & 0.090\,[1.17] & 0.168\,[1.14] & 0.052\,[1.39] & 0.090\,[1.42] & 0.149\,[1.59] \\
 & \rwr{} & \textbf{0.077} & \textbf{0.147} & 0.037 & \textbf{0.063} & \textbf{0.094} \\
\addlinespace[2pt]
$u_z$ & Transolver & 0.080\,[1.11] & 0.207\,[1.27] & \textbf{0.060}\,[0.72] & \textbf{0.086}\,[0.94] & 0.150\,[1.20] \\
 & GeoTransolver & 0.104\,[1.45] & 0.286\,[1.75] & 0.072\,[0.87] & 0.092\,[1.01] & 0.136\,[1.08] \\
 & \rwp{} & 0.091\,[1.27] & 0.218\,[1.33] & 0.092\,[1.11] & 0.121\,[1.32] & 0.199\,[1.59] \\
 & \rwr{} & \textbf{0.072} & \textbf{0.164} & 0.083 & 0.092 & \textbf{0.125} \\
\bottomrule
\end{tabular}}
\caption{Held-out test errors on the branched pipe at matched training budget for all output fields: relative $L_2$ error of each field and of its first spatial gradients, and band-resolved fRMSE over the low, mid, and high wavenumber bands. Bracketed values give the ratio of each model's error to \rwr{}'s; bold marks the best value per metric and field. The pressure results are discussed in App.~\ref{app:pressure}.}
\label{tab:full-metrics}
\end{table*}

\subsection{Composition Ablation}
\label{app:composition}

A stacked composition that runs all relaxation sweeps before or after the latent-attention iterations contains the same two mechanisms as an interleaved one. Table~\ref{tab:interleaving} compares four allocations of an identical budget of eight relaxation sweeps and four latent-attention iterations, from the two stacked orderings to progressively finer interleavings.

\begin{table}[htbp]
\centering
\small
\begin{tabular}{@{}lccc@{}}
\toprule
Steps & $p$ & $u_x$ & $u_z$ \\
\midrule
$(8,4,0){\times}1$ & 5.8 & 5.8 & 4.8 \\
$(0,4,8){\times}1$ & 6.0 & 5.3 & 4.7 \\
$(2,2,2){\times}2$ & 5.2 & 5.2 & 4.5 \\
$(1,1,1){\times}4$ & 5.1 & 5.1 & 4.5 \\
\bottomrule
\end{tabular}
\caption{Composition ablation on the branched pipe at a fixed budget of eight relaxation sweeps and four latent iterations: held-out test normalized mean absolute error (NMAE, $\times 10^{-2}$) of pressure and the two velocity components. A configuration $(m_1,r,m_2){\times}n_B$ runs $m_1$ relaxation sweeps, $r$ latent iterations, and $m_2$ further sweeps per block, repeated over $n_B$ blocks: $(8,4,0)$ places all relaxation before the latent path, $(0,4,8)$ after it, and the remaining rows interleave the two at increasing granularity.}
\label{tab:interleaving}
\end{table}

\subsection{Wall Diagnostics}
\label{app:wall}

\begin{figure*}[htbp]
\centering
\includegraphics[width=0.88\linewidth]{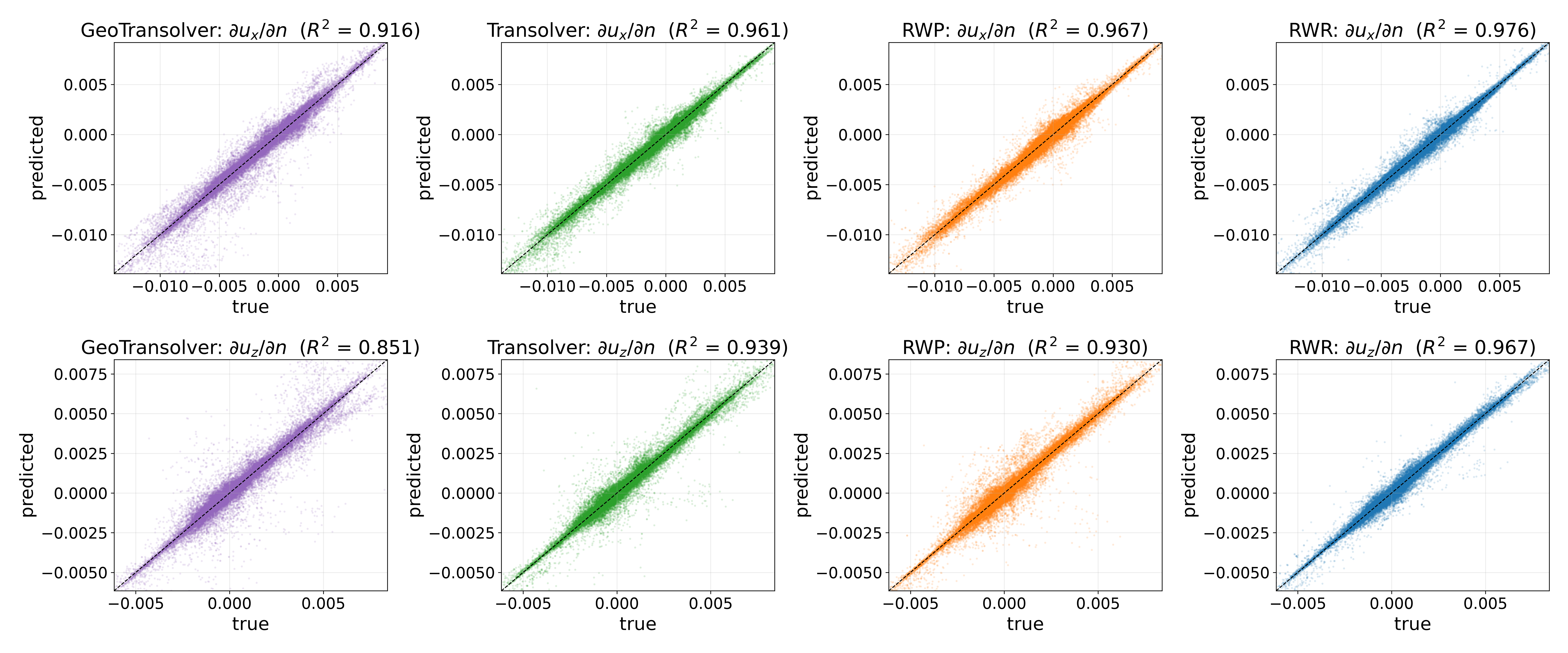}
\caption{Parity plots of the wall-normal derivatives $\partial u_x/\partial n$ and $\partial u_z/\partial n$ over wall cells on held-out branched pipe cases, with the coefficient of determination $R^2$ per model.}
\label{fig:wall-parity}
\end{figure*}

\begin{figure*}[htbp]
\centering
\includegraphics[width=0.88\linewidth]{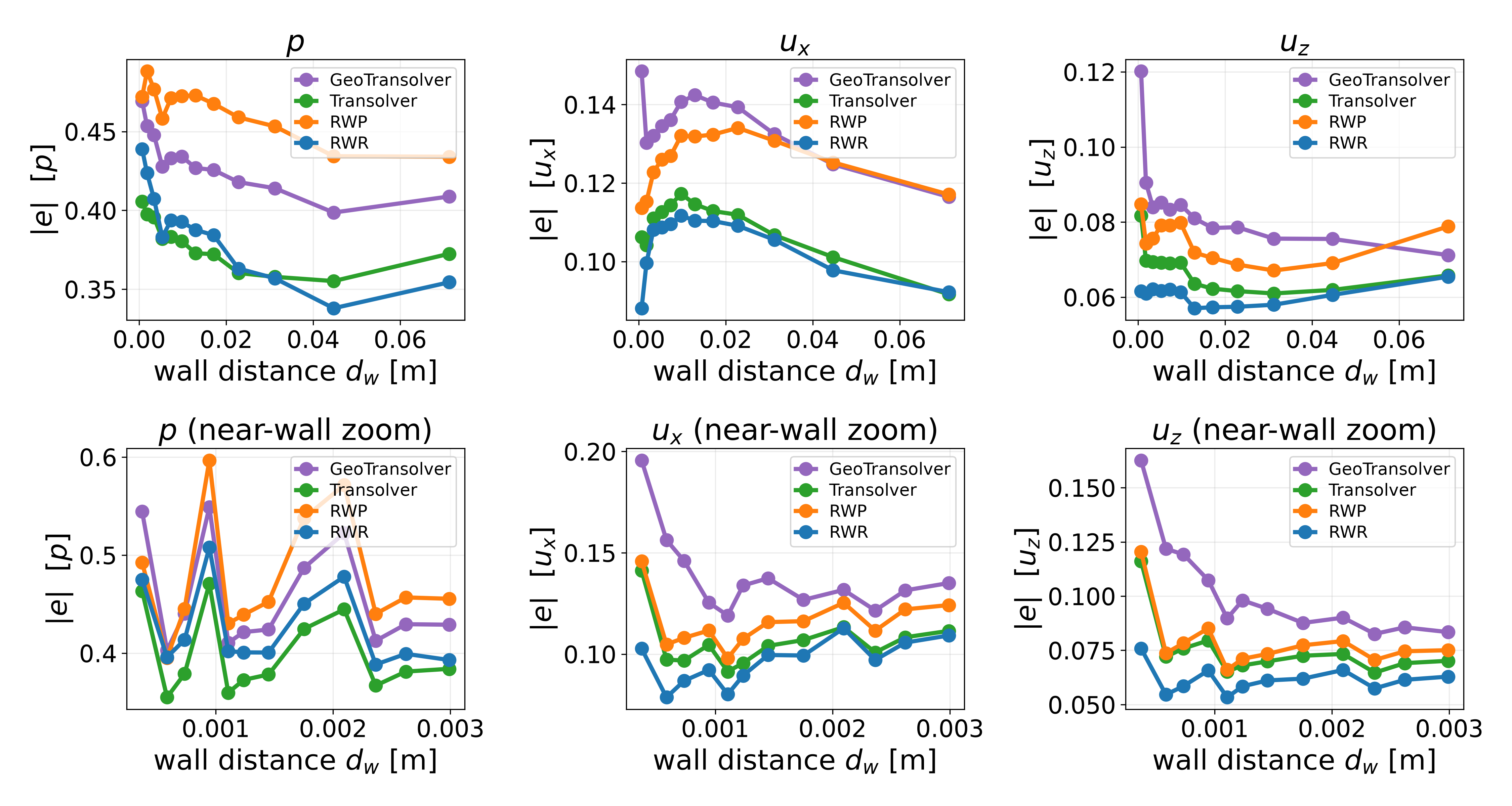}
\caption{Mean absolute error of pressure and the velocity components as a function of wall distance $d_w$ on held-out branched pipe cases (top row), with a near-wall zoom (bottom row).}
\label{fig:wall-distance}
\end{figure*}

Fig.~\ref{fig:wall-parity} shows parity plots of the wall-normal velocity derivatives over wall cells, and Fig.~\ref{fig:wall-distance} the mean absolute error of each field as a function of wall distance, with a near-wall zoom.

\subsection{Analysis Training Details}
\label{app:infra}

All analysis models are trained on the branched pipe in the same pipeline, under the 100-sample branched pipe training protocol of the main benchmarks (App.~\ref{app:internal-benchmark}). MGN, which is absent from the benchmarks, trains with an initial learning rate of $3 \times 10^{-4}$ for the same 550 epochs as \rwp{} and \rwr{}. Table~\ref{tab:rwp-rwr-config} lists the architectural configuration of \rwp{} and \rwr{}. The latent queries are geometry-anchored with learned anchor coordinates, passed through the shared random-Fourier-feature embedding ($\sigma = 0.5$, learnable coefficients, dimension equal to the hidden width), and the boundary-only read is enabled in all runs. MGN runs 15 message-passing sweeps at hidden width 128. Transolver is our reimplementation with 8 layers, hidden width 172, 4 attention heads, 32 slices, and MLP ratio 2; GeoTransolver is our reimplementation with 4 layers, hidden width 144, 8 heads, 32 slices, and ball-query radii $\{0.25, 1.0\}$ on normalized coordinates with $\{8, 32\}$ neighbors.

\begin{table}[htbp]
\centering
\small
\begin{tabular}{@{}lcc@{}}
\toprule
& \rwp{} & \rwr{} \\
\midrule
steps $(m_1, r, m_2) \times n_B$ & $(0,2,0) \times 2$ & $(2,2,2) \times 2$ \\
latents $L$ & 128 & 128 \\
hidden width & 192 & 128 \\
attention heads & 4 & 4 \\
latent self-attention & none & none \\
boundary-only read & yes & yes \\
weight sharing across $r$ & no & no \\
\bottomrule
\end{tabular}
\caption{Architectural configuration of \rwp{} and \rwr{} for the branched pipe analysis. Hidden widths are chosen to approximately match parameter counts across MGN, \rwp{}, and \rwr{}.}
\label{tab:rwp-rwr-config}
\end{table}

\section{Main Benchmark Results}
\label{app:internal-benchmark}

Table~\ref{tab:internal-full} reports the per-field NMAE, parameter counts, peak memory, and training cost behind Table~\ref{tab:internal}.

\paragraph{Training protocol}
All models are trained using the same pipeline under the same training protocol. Each dataset contains $300$ simulations, split $70/15/15$ into train/validation/test ($210/45/45$ samples). The data-limited variants use only the first $50$ or $100$ samples, split the same way ($35/7/8$ and $70/15/15$). 
Inputs and outputs are cell-centered mesh fields for the branched pipe data set and node-centered for the pump data sets. Input fields are standardized per component, with node coordinates appended. 
We minimize the mean-squared error with AdamW (batch size $1$, gradient-norm clipping at $2.0$) under a linear schedule that decays the learning rate to one-tenth of its base value over the first ($n_{\text{epoch}}-100$) epochs and holds it constant thereafter. The base learning rate is $10^{-3}$ for the branched pipe and $3\times10^{-4}$ for the pump. 
The training budget on the branched pipe is $550$ epochs for \rwp{} and \rwr{} and $650$ epochs for the attention baselines. For the pump-casing- and full-pump experiments all models are trained for $600$ and $500$ epochs respectively.
For all completed runs we report metrics computed on the test set. Each run uses a single A100 80\,GB GPU with gradient checkpointing enabled.

\paragraph{Model configurations}
\rwr{} and \rwp{} share a MeshGraphNet backbone (two blocks with hidden width $128$ on the branched pipe and casing, and width $96$ on the full pump) wrapping a latent-attention bottleneck: 
\rwr{} interleaves message-passing sweeps with the latent read/write ($\texttt{steps}=[2,2,2]$), while \rwp{} performs the read/write alone ($\texttt{steps}=[0,2,0]$). 
Transolver consists of eight layers with hidden width $256$ and uses eight heads.
GeoTransolver consists of four layers with width $256$, eight heads, and local geometric features gathered within radii $\{0.25,1.0\}$. 
Exact parameter counts per configuration are listed in Table~\ref{tab:internal-full}.

\paragraph{Sweeps and selection}
For every model family we sweep its principal capacity- and locality/geometry-governing hyperparameters and report the best model variant in Table~\ref{tab:internal-best-models} and their corresponding performance metrics in Table~\ref{tab:internal-full}. 
For \rwr{} and \rwp{} we vary the processor depth (two or three blocks), the latent-token count $L$, and whether the read attends to the full volume or only the boundary surface. 
Transolver and GeoTransolver are swept over their slice count and, in the data-limited setting, their hidden width.

\begin{table*}[htbp]
\centering
\footnotesize
\setlength{\tabcolsep}{2.5pt}
\caption{Architectural details of best performing models per family for each mesh and data setting.}
\label{tab:internal-best-models}
\resizebox{\linewidth}{!}{%
\begin{tabular}{l ccc ccc ccccc ccccc}
\toprule
& \multicolumn{3}{c}{Transolver} & \multicolumn{3}{c}{GeoTransolver} & \multicolumn{5}{c}{\rwr{}} & \multicolumn{5}{c}{\rwp{}}\\
\cmidrule(lr){2-4}\cmidrule(lr){5-7}\cmidrule(lr){8-12}\cmidrule(lr){13-17}
\#samples & \#params & \#slices & width & \#params & \#slices & width & \#params & \#latents & blocks & read & width & \#params & \#latents & blocks & read & width\\
\midrule
\multicolumn{17}{l}{\textbf{Branched pipe ($\mathbf{{\sim}60}$k cells)}}\\
\phantom{0}50 & 3.58M & 32 & 172 & 4.02M & 32 & 144 & 3.58M & \phantom{0}64 & 2 & bdy & 128 & 3.87M & 512 & 2 & bdy & 128\\
100 & 3.58M & 32 & 172 & 4.02M & 16 & 144 & 2.79M & 128 & 2 & bdy & 128 & 5.65M & 128 & 3 & bdy & 128\\
300 & 7.72M & 32 & 256 & 9.53M & 32 & 256 & 5.30M & 128 & 3 & bdy & 128 & 2.52M & 128 & 3 & bdy & 128\\
\midrule
\multicolumn{17}{l}{\textbf{Pump casing ($\mathbf{{\sim}270}$k cells)}}\\
\phantom{0}50 & 7.89M & 32 & 256 & 9.53M & 32 & 256 & 5.30M & 128 & 3 & bdy & 128 & 5.65M & 128 & 3 & bdy & 128\\
100 & 7.89M & 32 & 256 & 9.53M & 32 & 256 & 5.30M & 128 & 3 & bdy & 128 & 5.65M & 128 & 3 & bdy & 128\\
300 & 7.72M & 32 & 256 & 9.52M & 16 & 256 & 5.30M & 128 & 3 & bdy & 128 & 2.52M & 128 & 3 & bdy & 128\\
\midrule
\multicolumn{17}{l}{\textbf{Pump full ($\mathbf{{\sim}1.4}$M cells)}}\\
\phantom{0}50 & 1.10M & 32 & \phantom{0}96 & 2.46M & 32 & \phantom{0}96 & 2.03M & 512 & 2 & bdy & 128 & 1.73M & 128 & 2 & bdy & 128\\
100 & 1.10M & 32 & \phantom{0}96 & 2.46M & 32 & \phantom{0}96 & 2.02M & 128 & 2 & bdy & 128 & 1.73M & 128 & 2 & bdy & 128\\
300 & 1.10M & 32 & \phantom{0}96 & 1.99M & 32 & \phantom{0}80 & 2.03M & 512 & 2 & bdy & \phantom{0}96 & 1.73M & 128 & 2 & all & 128\\
\bottomrule
\end{tabular}}
\end{table*}

\begin{table*}[htbp]
\centering
\footnotesize
\setlength{\tabcolsep}{5pt}
\renewcommand{\arraystretch}{1.0}
\caption{Performance metrics of best models per family for each mesh and data setting. 
We report the NMAE per output field, across-field mean NMAE, parameter count, peak memory, average wall-clock time per epoch, and the
number of epochs to reach a mean $R^2\geq0.95$ (n/r: not reached). 
The branched pipe is quasi-2D, so $u_y$ is absent.
}
\label{tab:internal-full}
\begin{tabular}{c ccccc c c c c}
\toprule
& \multicolumn{5}{c}{NMAE ($\times10^{-2}$)}\\
\cmidrule(lr){2-6}
    Model & $p$ & $u_x$ & $u_y$ & $u_z$ & mean & \#params & \makecell{peak\\ memory} & \makecell{average\\ time/epoch} & \makecell{\#epochs until\\ avg. $R^2\geq0.95$}\\
\midrule
\multicolumn{10}{l}{\textbf{Branched pipe ($\mathbf{{\sim}60}$k cells)}}\\
\multicolumn{10}{l}{50 samples}\\
Transolver & 10.3 & 12.7 &  & 10.8 & 11.3 & 3.58M & \phantom{0}8.1 GB & 8s & 470\\
GeoTransolver & 9.5 & 11.8 &  & 10.2 & 10.5 & 4.02M & \phantom{0}8.0 GB & 7s & 370\\
\rwp{} & 10.4 & 10.7 &  & 9.7 & 10.3 & 3.87M & \phantom{0}4.2 GB & 6s & 290\\
\rwr{} & \textbf{8.7} & \textbf{8.9} &  & \textbf{7.7} & \textbf{8.4} & 3.58M & 13.5 GB & 11s & 160\\
\multicolumn{10}{l}{100 samples}\\
Transolver & \textbf{8.5} & 7.1 &  & 7.0 & 7.5 & 3.58M & \phantom{0}6.9 GB & 16s & 220\\
GeoTransolver & 9.6 & 7.8 &  & 7.5 & 8.3 & 4.02M & \phantom{0}7.4 GB & 14s & 190\\
\rwp{} & 8.7 & 6.7 &  & 6.4 & 7.3 & 5.65M & \phantom{0}4.6 GB & 15s & 140\\
\rwr{} & 8.7 & \textbf{6.3} &  & \textbf{5.8} & \textbf{6.9} & 2.79M & 10.1 GB & 23s & 110\\
\multicolumn{10}{l}{300 samples}\\
Transolver & \textbf{4.8} & 5.3 &  & 4.5 & 4.8 & 7.72M & 13.5 GB & 53s & 110\\
GeoTransolver & 5.9 & 5.6 &  & 5.0 & 5.5 & 9.53M & 15.6 GB & 51s & 140\\
\rwp{} & 5.0 & 5.4 &  & 4.5 & 5.0 & 2.52M & \phantom{0}3.6 GB & 40s & \phantom{0}50\\
\rwr{} & 5.0 & \textbf{5.0} &  & \textbf{4.3} & \textbf{4.7} & 5.30M & 26.7 GB & 95s & \phantom{0}40\\
\midrule
\multicolumn{10}{l}{\textbf{Pump casing ($\mathbf{{\sim}270}$k cells)}}\\
\multicolumn{10}{l}{50 samples}\\
Transolver & \phantom{0}6.4 & 12.0 & 30.9 & 13.0 & 15.6 & 7.89M & 28.5 GB & 44s & 230\\
GeoTransolver & \phantom{0}4.6 & 9.3 & 24.5 & 10.0 & 12.1 & 9.53M & 31.8 GB & 43s & 160\\
\rwp{} & \phantom{0}\textbf{3.9} & \textbf{8.8} & 23.5 & 9.5 & 11.4 & 5.65M & 14.2 GB & 30s & \phantom{0}80\\
\rwr{} & \phantom{0}4.3 & 8.8 & \textbf{21.0} & \textbf{9.1} & \textbf{10.8} & 5.30M & 21.8 GB & 57s & \phantom{0}70\\
\multicolumn{10}{l}{100 samples}\\
Transolver & \phantom{0}3.9 & 8.2 & 23.1 & 8.3 & 10.9 & 7.89M & 28.6 GB & 87s & 120\\
GeoTransolver & \phantom{0}3.1 & 7.0 & 20.0 & 7.1 & 9.3 & 9.53M & 31.8 GB & 85s & \phantom{0}80\\
\rwp{} & \phantom{0}\textbf{2.6} & 6.3 & 17.2 & \textbf{6.3} & 8.1 & 5.65M & 14.1 GB & 59s & \phantom{0}30\\
\rwr{} & \phantom{0}2.6 & \textbf{6.1} & \textbf{15.5} & 6.3 & \textbf{7.7} & 5.30M & 21.9 GB & 111s & \phantom{0}40\\
\multicolumn{10}{l}{300 samples}\\
Transolver & \phantom{0}2.5 & 6.0 & 16.8 & 6.1 & 7.9 & 7.72M & 31.4 GB & 265s & \phantom{0}40\\
GeoTransolver & \phantom{0}2.2 & 5.3 & 14.7 & 5.3 & 6.9 & 9.52M & 29.6 GB & 241s & \phantom{0}40\\
\rwp{} & \phantom{0}2.1 & 5.1 & 13.0 & 5.2 & 6.3 & 2.52M & \phantom{0}9.9 GB & 156s & \phantom{0}20\\
\rwr{} & \phantom{0}\textbf{1.7} & \textbf{4.5} & \textbf{10.9} & \textbf{4.5} & \textbf{5.4} & 5.30M & 22.0 GB & 331s & \phantom{0}20\\
\midrule
\multicolumn{10}{l}{\textbf{Pump full ($\mathbf{{\sim}1.4}$M cells)}}\\
\multicolumn{10}{l}{50 samples}\\
Transolver & 33.0 & 27.4 & 62.8 & 25.1 & 37.1 & 1.10M & 75.2 GB & 170s & n/r\\
GeoTransolver & 26.8 & 24.7 & 54.9 & 22.4 & 32.2 & 2.46M & 79.2 GB & 148s & n/r\\
\rwp{} & \phantom{0}9.1 & 19.8 & 40.2 & 18.0 & 21.8 & 1.73M & 33.7 GB & 91s & n/r\\
\rwr{} & \phantom{0}\textbf{7.8} & \textbf{17.3} & \textbf{36.0} & \textbf{15.9} & \textbf{19.3} & 2.03M & 59.4 GB & 182s & n/r\\
\multicolumn{10}{l}{100 samples}\\
Transolver & \phantom{0}9.1 & 21.3 & 43.2 & 19.1 & 23.2 & 1.10M & 75.5 GB & 331s & n/r\\
GeoTransolver & — & — & — & — & — & 2.46M & 79.2 GB & 285s & n/r\\
\rwp{} & \phantom{0}6.8 & 15.9 & 31.7 & 15.0 & 17.3 & 1.73M & 34.3 GB & 177s & 200\\
\rwr{} & \phantom{0}\textbf{6.4} & \textbf{13.4} & \textbf{28.5} & \textbf{12.5} & \textbf{15.2} & 2.02M & 59.1 GB & 340s & 140\\
\multicolumn{10}{l}{300 samples}\\
Transolver & \phantom{0}5.9 & 14.3 & 30.1 & 13.4 & 15.9 & 1.10M & 75.7 GB & 986s & 160\\
GeoTransolver & \phantom{0}\textbf{5.3} & 12.3 & 27.6 & 11.4 & 14.1 & 1.99M & 73.8 GB & 830s & 170\\
\rwp{} & \phantom{0}5.4 & 12.7 & 27.5 & 11.9 & 14.4 & 1.73M & 44.9 GB & 968s & 100\\
\rwr{} & \phantom{0}5.7 & \textbf{10.8} & \textbf{24.2} & \textbf{9.9} & \textbf{12.7} & 2.03M & 59.6 GB & 985s & \phantom{0}60\\
\bottomrule
\end{tabular}
\end{table*}

\section{Additional Benchmarks}
\label{app:public}
\paragraph{Datasets}
AirfRANS: two-dimensional RANS airfoil flows with
roughly 180k nodes per case, velocity, pressure, and turbulent viscosity as
targets, and per-case inlet velocity and angle of attack as global
parameters~\cite{bonnet2022airfrans}. AhmedML: three-dimensional surface
meshes of roughly 1.1M points over 500 Ahmed-body variants, surface pressure
and wall shear stress as targets, eight global geometry parameters, and a
400/50/50 split~\cite{ashton2024ahmedml}. Geo-FNO airfoil and
pipe~\cite{li2023geofno} and Darcy flow~\cite{li2021fourier}, with the standard
splits of roughly 1000 training and 200 test cases used by published
Transolver-family reproductions.

\paragraph{Input parity} Every model receives the same per-case information. The MGN-family models encode geometry through relative edge displacements, \rwr{} additionally through the positional embeddings of its global stage, and Transolver and GeoTransolver receive coordinates concatenated per token. Per-case global parameters are injected into every message-passing step for the MGN-family models, broadcast-concatenated per token for \rwp{} and Transolver, and passed through the dedicated global-context tokenizer for GeoTransolver. The read attends to all nodes on every dataset; on AhmedML the boundary-only read is enabled but coincides with the full read, since every node lies on the surface.

\paragraph{Model Hyperparameters}
\rwr{} uses 3 blocks on AirfRANS and AhmedML and 2 blocks on the Geo-FNO datasets; \rwp{} uses 4 and 3 blocks, respectively. For the read-write stage, we use 128 latents on AirfRANS and AhmedML and 64 latents on the Geo-FNO datasets, at a hidden width of 128 and with 4 attention heads throughout. Transolver and GeoTransolver are run with 8 layers, hidden width 256, MLP ratio 2, and 32 slices, and 4 layers with hidden width 256 and 32 slices, respectively~\cite{wu2024transolver,adams2025geotransolver}. GeoTransolver's multi-scale ball-query mechanism supports only three-dimensional coordinates and is active on the AhmedML surface dataset; on the two-dimensional datasets it runs without it. Our implementation of MeshTransolver uses a 2--2--2 structure of MGN--Transolver--MGN layers, with the two Transolver layers using 4 attention heads, 32 slices, and a hidden width of 128 on all datasets. All MGN-based models use the same 2-layer MLP layer widths of [128,128] on all datasets.

\paragraph{Training protocol}
AdamW with a base learning rate of $10^{-3}$ and weight decay 0.01, linear learning-rate decay to $0.1\times$ over the full run, and gradient-norm clipping at 2.0. Models train in bfloat16 on AirfRANS and AhmedML and in fp32 on the Geo-FNO datasets, with per-block activation checkpointing where required to fit memory. The batch
size is 1 on AirfRANS and AhmedML; on the Geo-FNO datasets it is 8 for the
MGN-family models and 1 for Transolver and GeoTransolver. Training runs 100 epochs on AirfRANS and AhmedML and 800 epochs on the Geo-FNO datasets; on the Geo-FNO airfoil each model reports the better of two settings, $(10^{-3}$, 800 epochs$)$ and $(5\times10^{-4}$, 1000 epochs$)$. The loss is the MSE on standardized fields, each run reports its best validation loss, and all results are single seed.

\section{Model Predictions}
\label{app:predictions}

\begin{figure*}[htbp]
\centering
\includegraphics[width=0.9\linewidth]{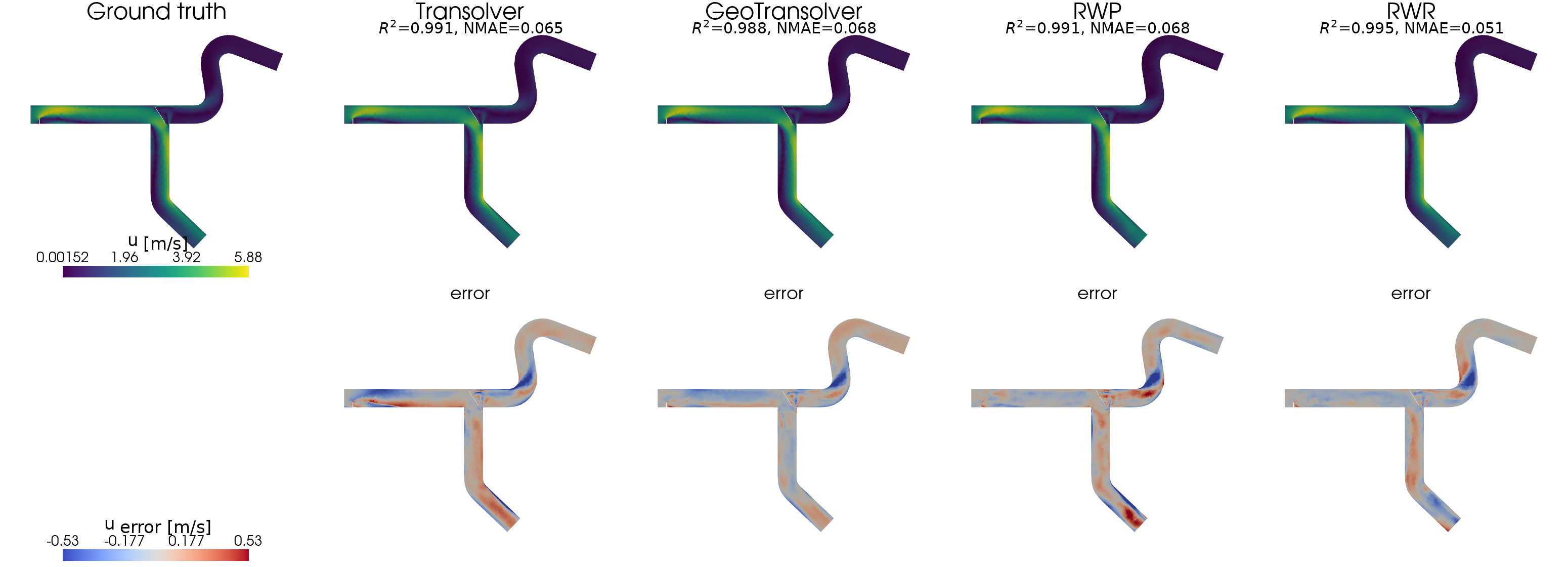}\\[4pt]
\includegraphics[width=0.9\linewidth]{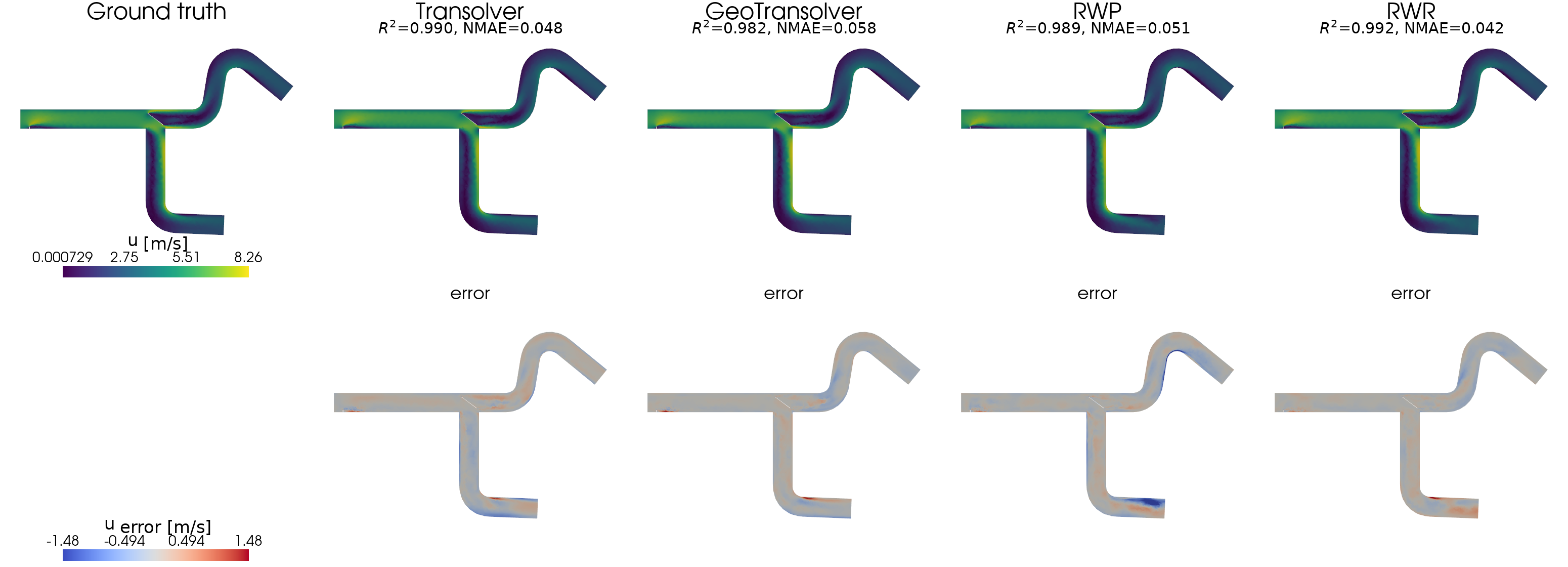}
\caption{Predicted velocity magnitude on the branched pipe for two held-out test cases (top and bottom groups), shown on the mid-$y$ slice. The left column shows the ground truth; each model column shows the prediction, with its $R^2$ and NMAE, above the signed pointwise error.}
\label{fig:pred-pipe-u}
\end{figure*}

\begin{figure*}[htbp]
\centering
\includegraphics[width=0.9\linewidth]{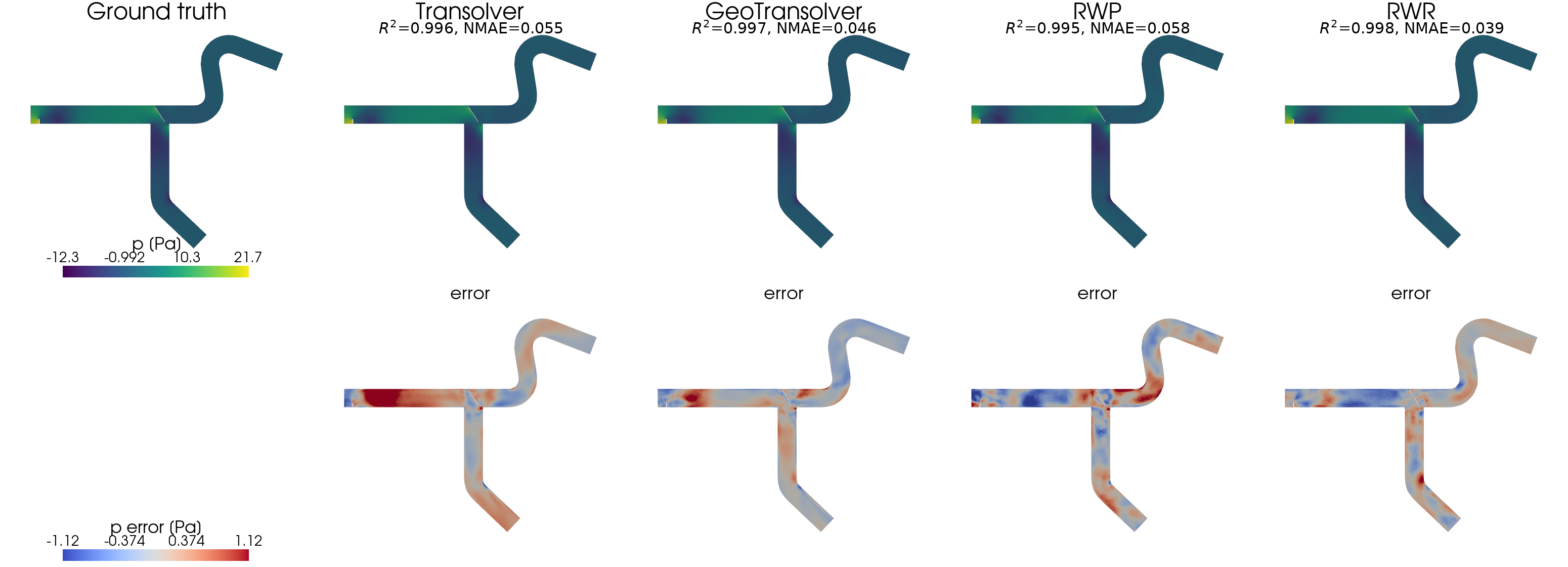}\\[4pt]
\includegraphics[width=0.9\linewidth]{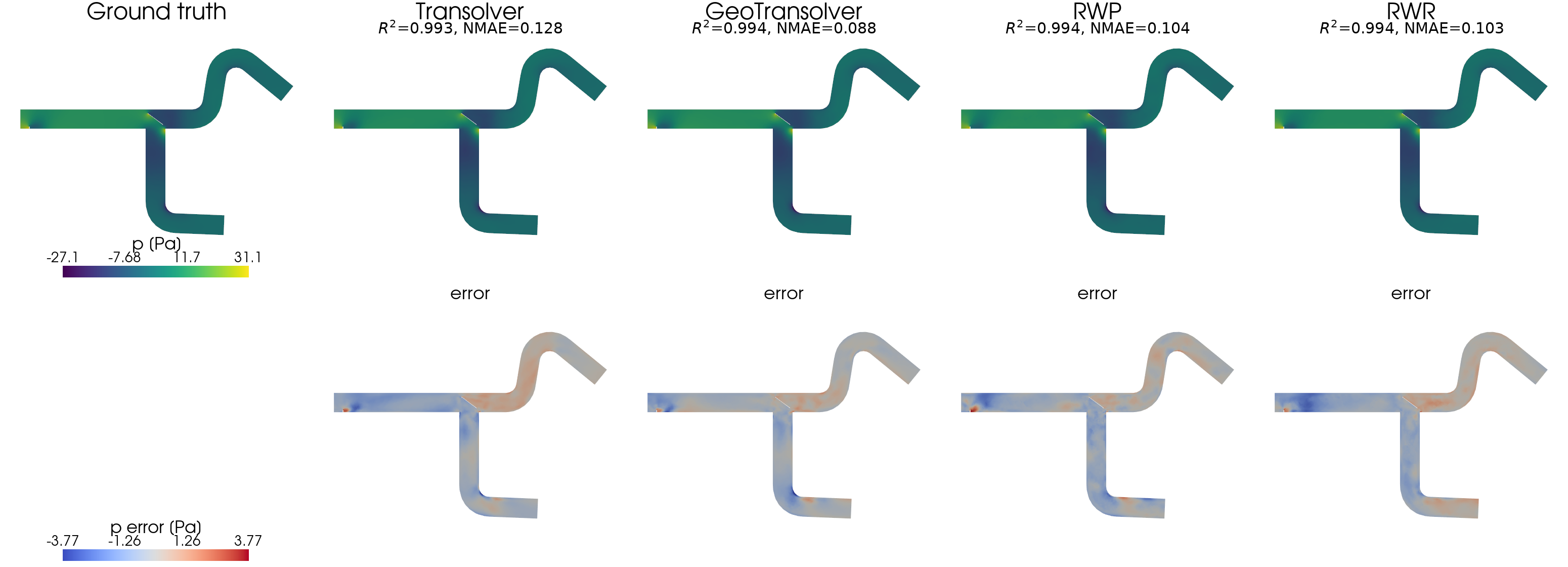}
\caption{Predicted pressure on the branched pipe for two held-out test cases (top and bottom groups), shown on the mid-$y$ slice. The left column shows the ground truth; each model column shows the prediction, with its $R^2$ and NMAE, above the signed pointwise error.}
\label{fig:pred-pipe-p}
\end{figure*}

\begin{figure*}[htbp]
\centering
\includegraphics[width=0.9\linewidth]{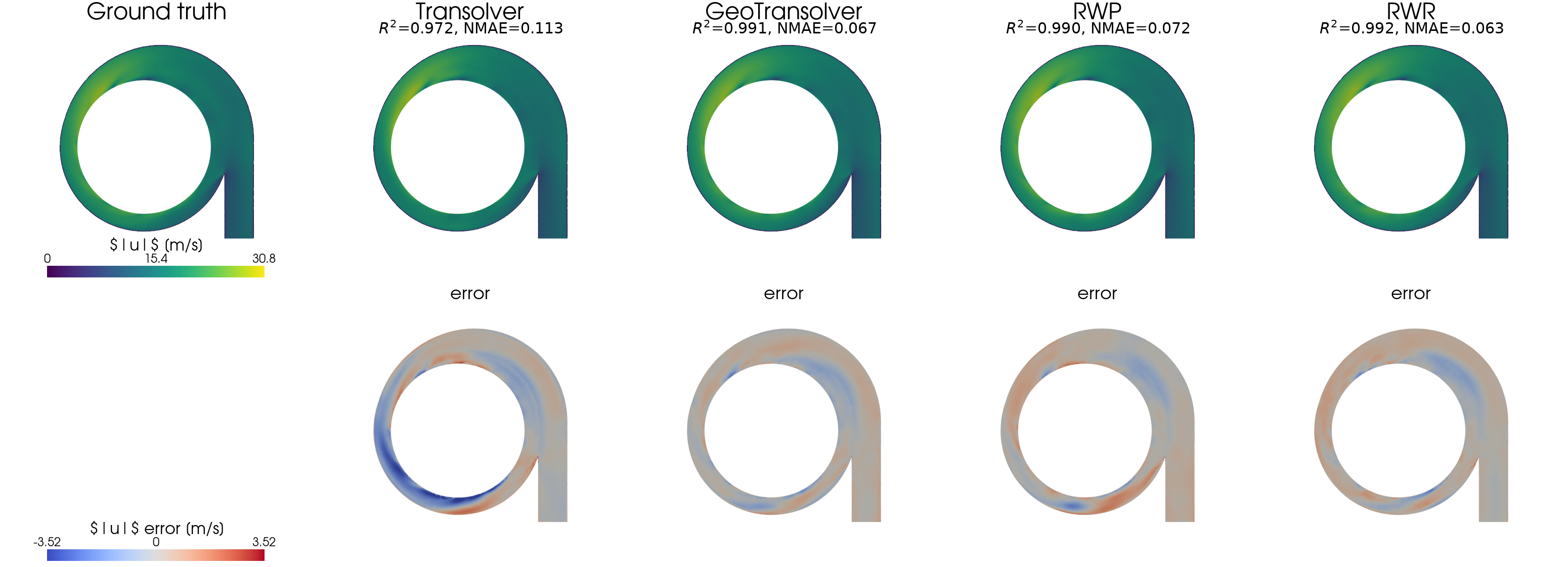}\\[4pt]
\includegraphics[width=0.9\linewidth]{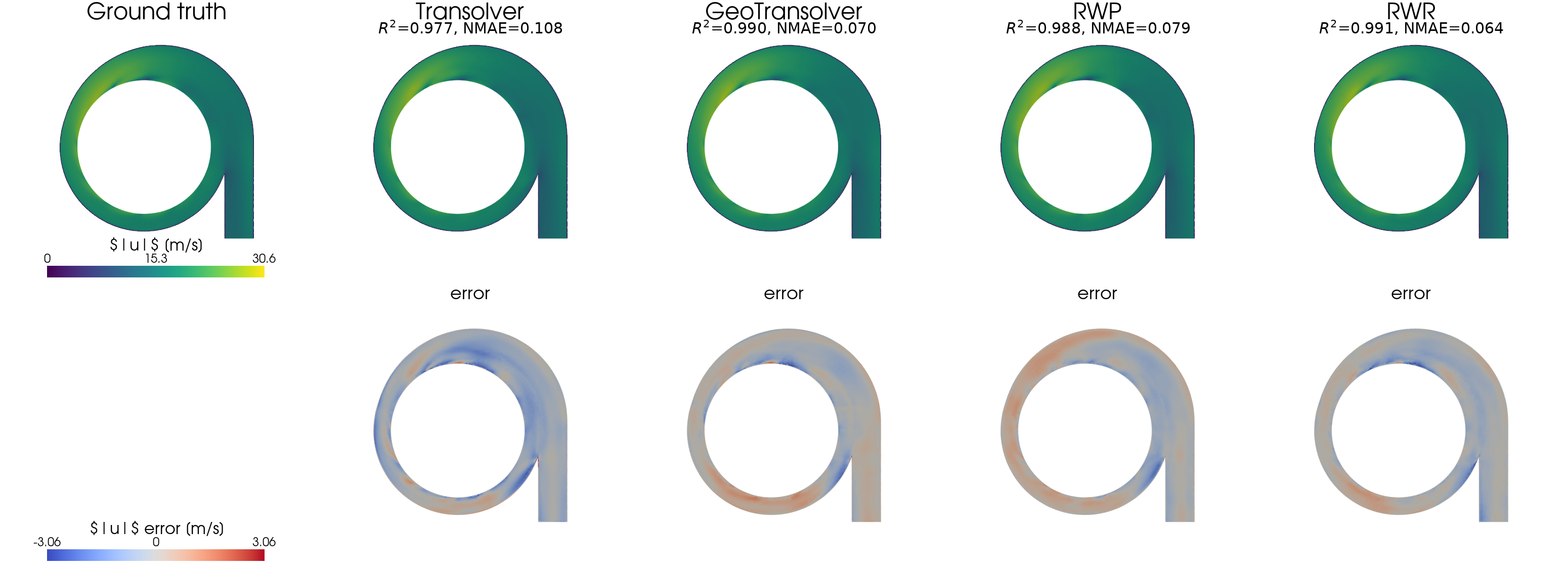}
\caption{Predicted velocity magnitude on the pump casing for two held-out test cases (top and bottom groups), shown on the $y = 0.04$\,m plane. The left column shows the ground truth; each model column shows the prediction, with its $R^2$ and NMAE, above the signed pointwise error.}
\label{fig:pred-casing-u}
\end{figure*}

\begin{figure*}[htbp]
\centering
\includegraphics[width=0.9\linewidth]{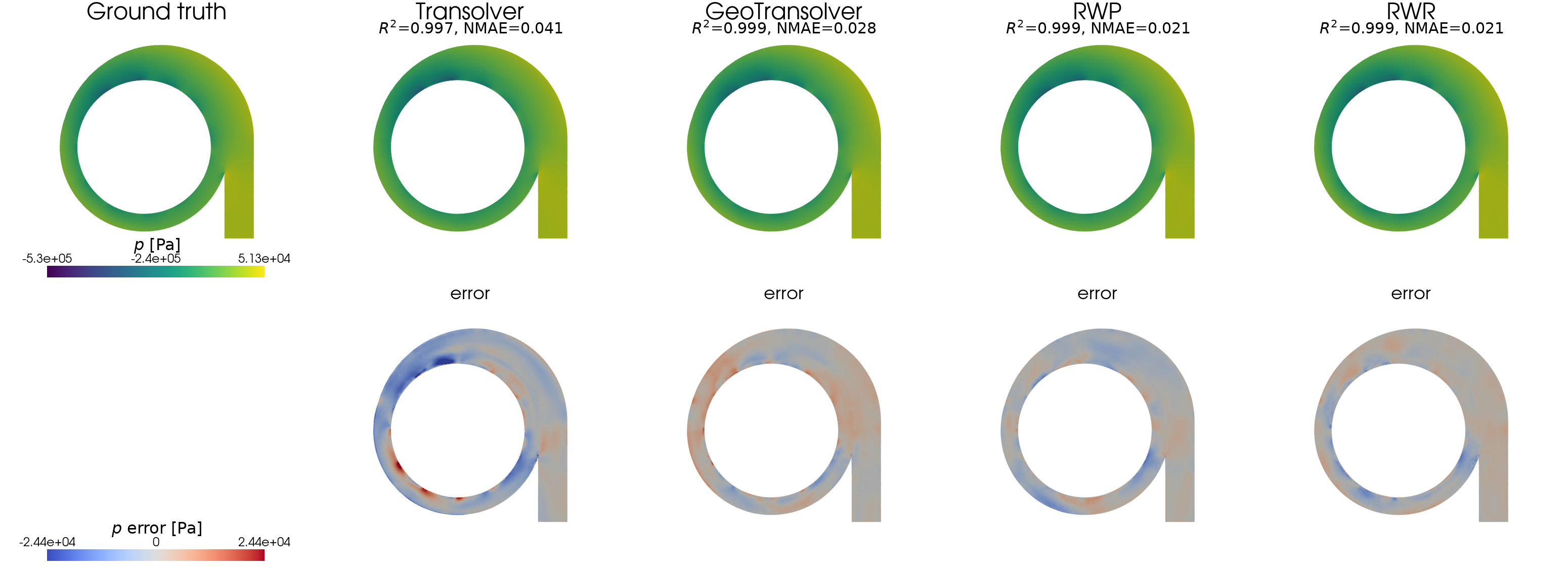}\\[4pt]
\includegraphics[width=0.9\linewidth]{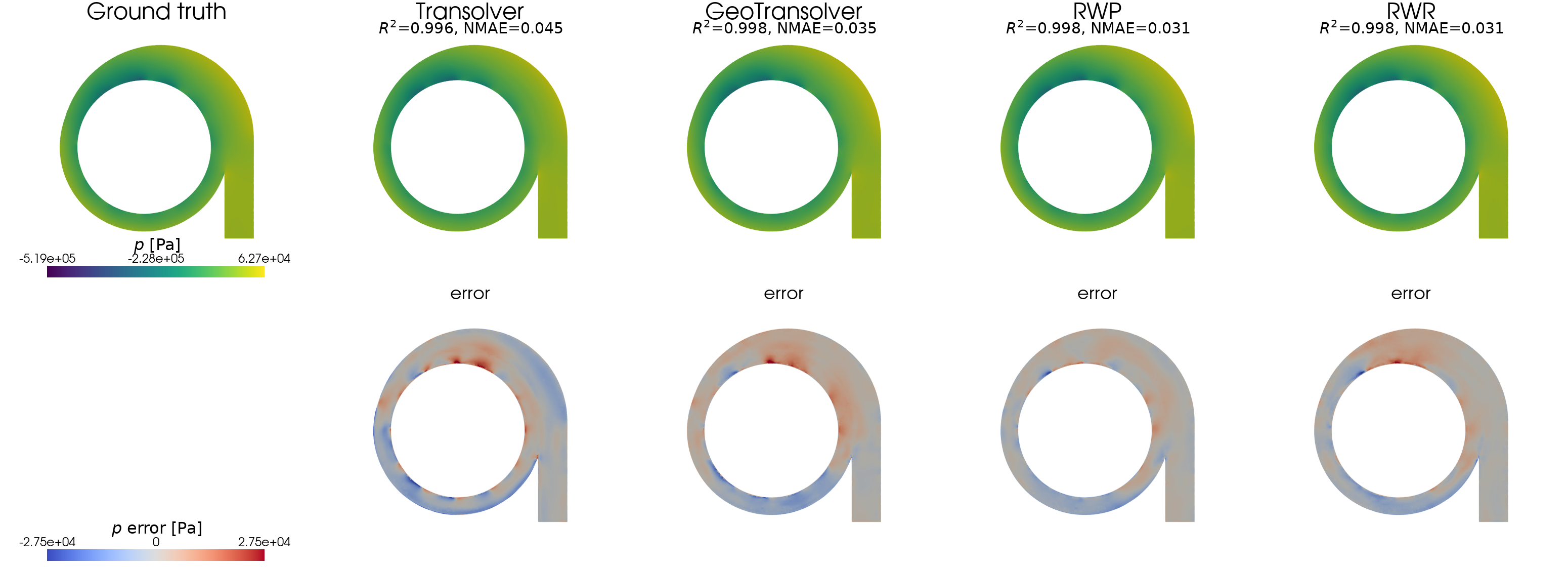}
\caption{Predicted pressure on the pump casing for two held-out test cases (top and bottom groups), shown on the $y = 0.04$\,m plane. The left column shows the ground truth; each model column shows the prediction, with its $R^2$ and NMAE, above the signed pointwise error.}
\label{fig:pred-casing-p}
\end{figure*}

\begin{figure*}[htbp]
\centering
\includegraphics[width=0.9\linewidth]{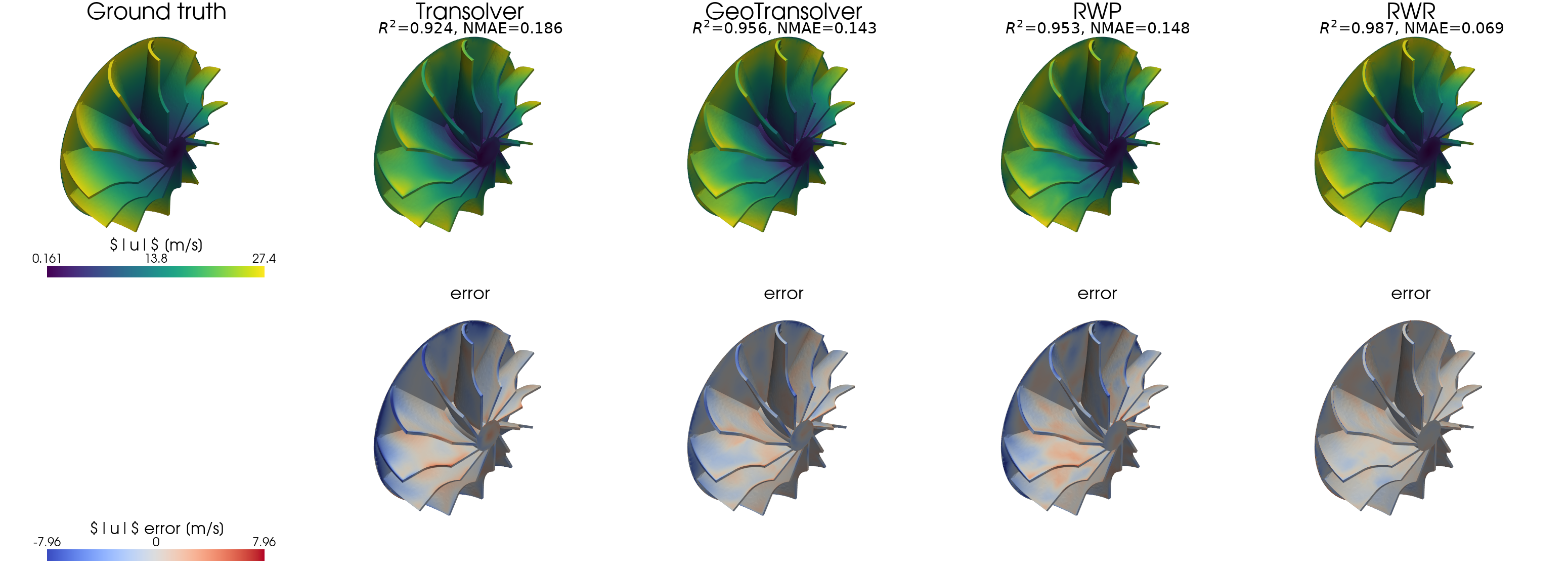}\\[4pt]
\includegraphics[width=0.9\linewidth]{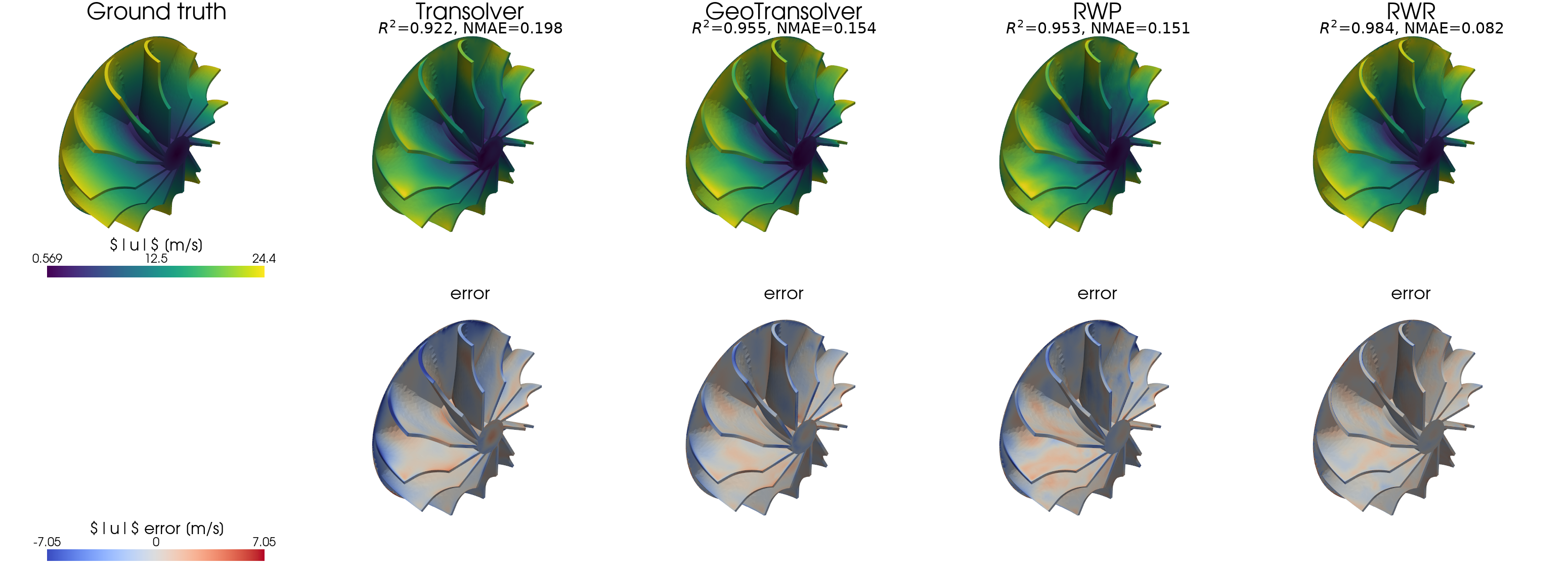}
\caption{Predicted velocity magnitude on the full pump mesh for two held-out test cases (top and bottom groups), shown on the impeller surface. The left column shows the ground truth; each model column shows the prediction, with its $R^2$ and NMAE, above the signed pointwise error.}
\label{fig:pred-full-u}
\end{figure*}

\begin{figure*}[htbp]
\centering
\includegraphics[width=0.9\linewidth]{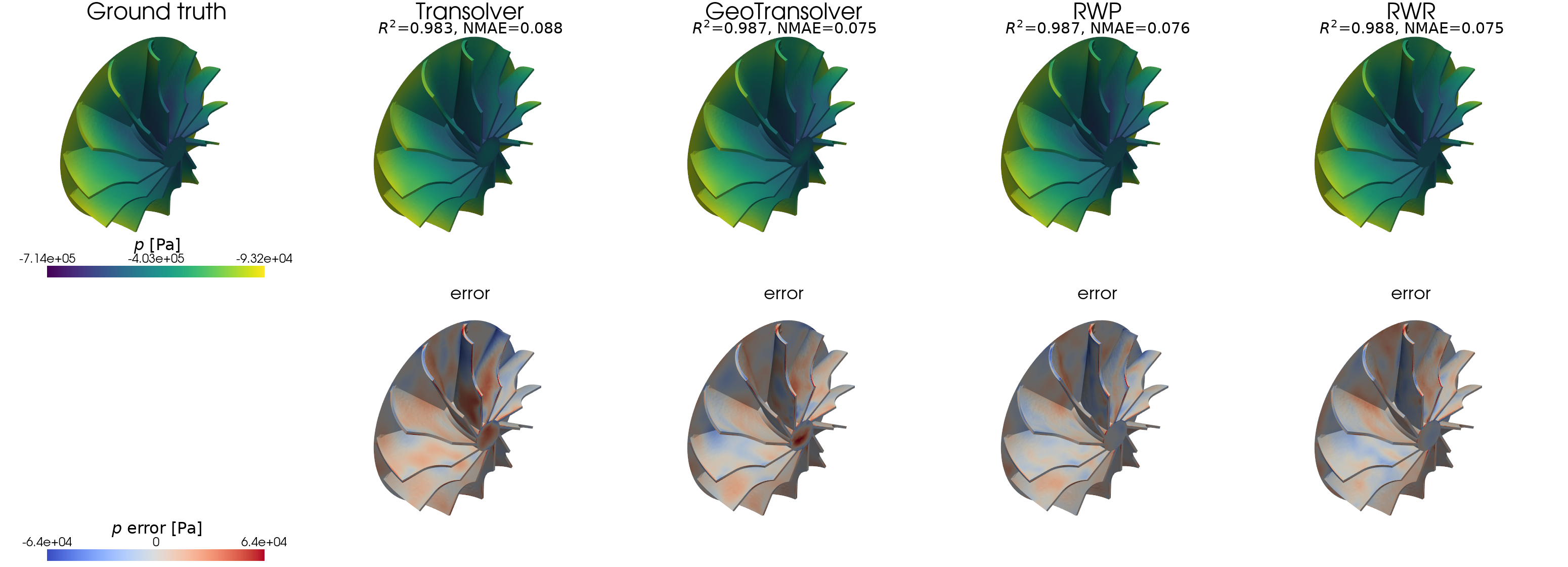}\\[4pt]
\includegraphics[width=0.9\linewidth]{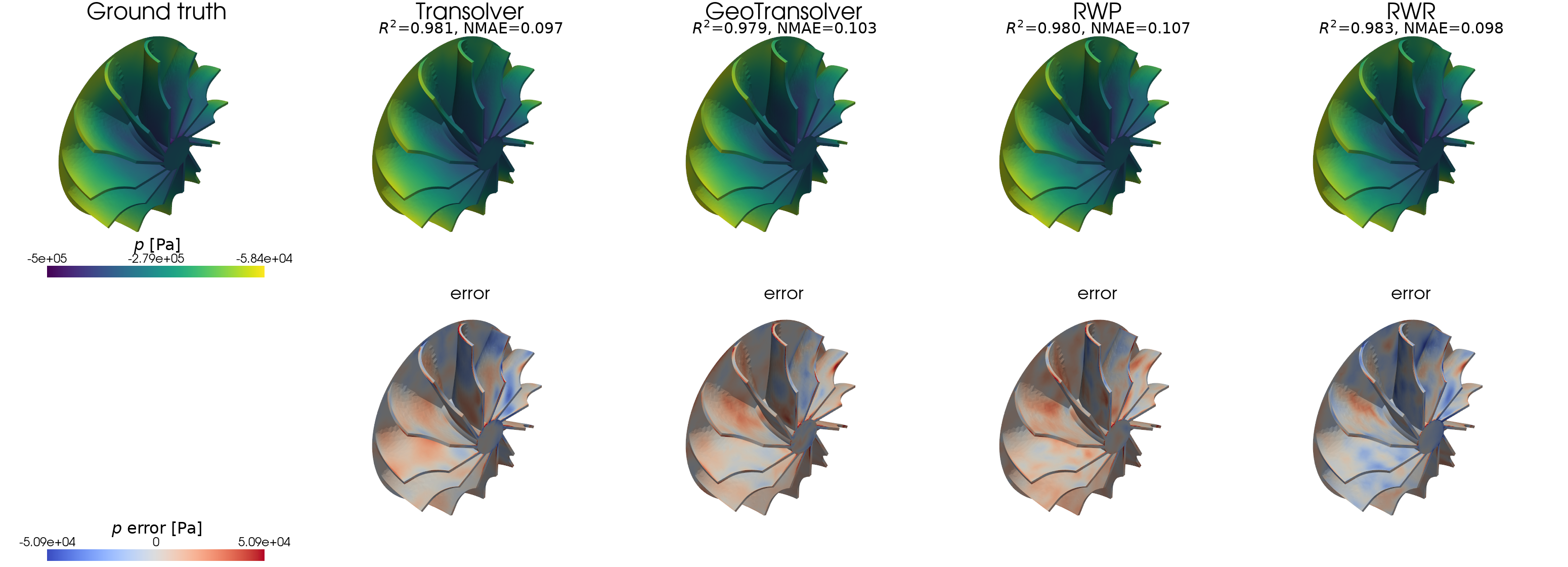}
\caption{Predicted pressure on the full pump mesh for two held-out test cases (top and bottom groups), shown on the impeller surface. The left column shows the ground truth; each model column shows the prediction, with its $R^2$ and NMAE, above the signed pointwise error.}
\label{fig:pred-full-p}
\end{figure*}

Fig.~\ref{fig:pred-pipe-u}--\ref{fig:pred-full-p} show model predictions and signed pointwise errors for two held-out test cases per dataset, for the pressure and the velocity magnitude. Branched pipe fields are shown on the mid-$y$ slice, pump-casing fields on the $y = 0.04$\,m plane, and full-pump fields on the impeller surface. Each model panel reports its $R^2$ and NMAE for the shown field and case.

\end{document}